\documentclass[10pt,journal,compsoc]{IEEEtran}

\ifCLASSOPTIONcompsoc
  \usepackage[nocompress]{cite}
\else
  \usepackage{cite}
\fi

\ifCLASSINFOpdf
\else
\fi

\usepackage{amsmath}
\usepackage{amssymb}
\usepackage{algorithmic}
\usepackage{array}
\usepackage{subfigure}
\usepackage{epsfig}
\usepackage{graphicx}
\usepackage{multirow}
\usepackage{color}
\usepackage{url}

\usepackage[framemethod=tikz]{mdframed}
\usepackage[colorlinks,linkcolor=blue]{hyperref}
\usepackage{fixltx2e}
\begin{document}
%
\title{Entity-enhanced Adaptive Reconstruction Network for Weakly Supervised Referring Expression Grounding}
%
%
%
%

\author{Xuejing Liu, Liang Li, Shuhui Wang, Zheng-Jun Zha, Zechao Li, Qi Tian,~\IEEEmembership{Fellow,~IEEE},\\  and Qingming Huang,~\IEEEmembership{Fellow,~IEEE}
\IEEEcompsocitemizethanks{
	\IEEEcompsocthanksitem X. Liu, L. Li, and S. Wang are with the Key Laboratory of Intelligent Information Processing of Chinese Academy of Sciences (CAS), Institute of Computing Technology, CAS, Beijing, 100190 China. X. Liu is also with University of Chinese Academy of Sciences, Beijing 100190, China. S. Wang is also with Peng Cheng Laboratory, Shenzhen 518066, China. E-mail: xuejing.liu@vipl.ict.ac.cn; \{liang.li, wangshuhui\}@ict.ac.cn.
	\IEEEcompsocthanksitem Z. Zha is with the School of Information Science and Technology, University of Science and Technology of China, Hefei 230027, China. E-mail: zhazj@ustc.edu.cn.
	\IEEEcompsocthanksitem Z. Li is with the School of Computer Science, Nanjing University of Science and Technology, Nanjing 210094, China. E-mail: zechao.li@njust.edu.cn.
	\IEEEcompsocthanksitem  Q. Tian is with Huawei Cloud \& AI, Shenzhen 518129, China. E-mail: wywqtian@gmail.com.
	\IEEEcompsocthanksitem Q. Huang is with the School of Computer and Control Engineering, University of Chinese Academy of Sciences, Beijing 100190, China, and also with the Institute of Computing Technology, CAS, Beijing 100190, China, and also with Peng Cheng Laboratory, Shenzhen 518066, China. 
	E-mail: qmhuang@ucas.ac.cn. 
	\IEEEcompsocthanksitem Corresponding author: Liang Li and Qingming Huang.
		}}

\markboth{Journal of \LaTeX\ Class Files,~Vol.~14, No.~8, August~2015}%
{Shell \MakeLowercase{\textit{et al.}}: Bare Demo of IEEEtran.cls for Computer Society Journals}

\IEEEtitleabstractindextext{%
\begin{abstract}
Weakly supervised Referring Expression Grounding (REG) aims to ground a particular target in an image described by a language expression while lacking the correspondence between target and expression. Two main problems exist in weakly supervised REG. First, the lack of region-level annotations introduces ambiguities between proposals and queries. Second, most previous weakly supervised REG methods ignore the discriminative location and context of the referent, causing difficulties in distinguishing the target from other same-category objects. To address the above challenges, we design an entity-enhanced adaptive reconstruction network (EARN). Specifically, EARN includes three modules: entity enhancement, adaptive grounding, and collaborative reconstruction. In entity enhancement, we calculate semantic similarity as supervision to select the candidate proposals.
Adaptive grounding calculates the ranking score of candidate proposals upon subject, location and context with hierarchical attention. Collaborative reconstruction measures the ranking result from three perspectives: adaptive reconstruction, language reconstruction and attribute classification. The adaptive mechanism helps to alleviate the variance of different referring expressions. Experiments on five datasets show EARN outperforms existing state-of-the-art methods. Qualitative results demonstrate that the proposed EARN can better handle the situation where multiple objects of a particular category are situated together.
\end{abstract}

%

}

\maketitle

\IEEEdisplaynontitleabstractindextext

%
\IEEEpeerreviewmaketitle

\IEEEraisesectionheading{\section{Introduction}\label{sec:introduction}}

\begin{figure}
	\centering
	\includegraphics[width=0.45\textwidth]{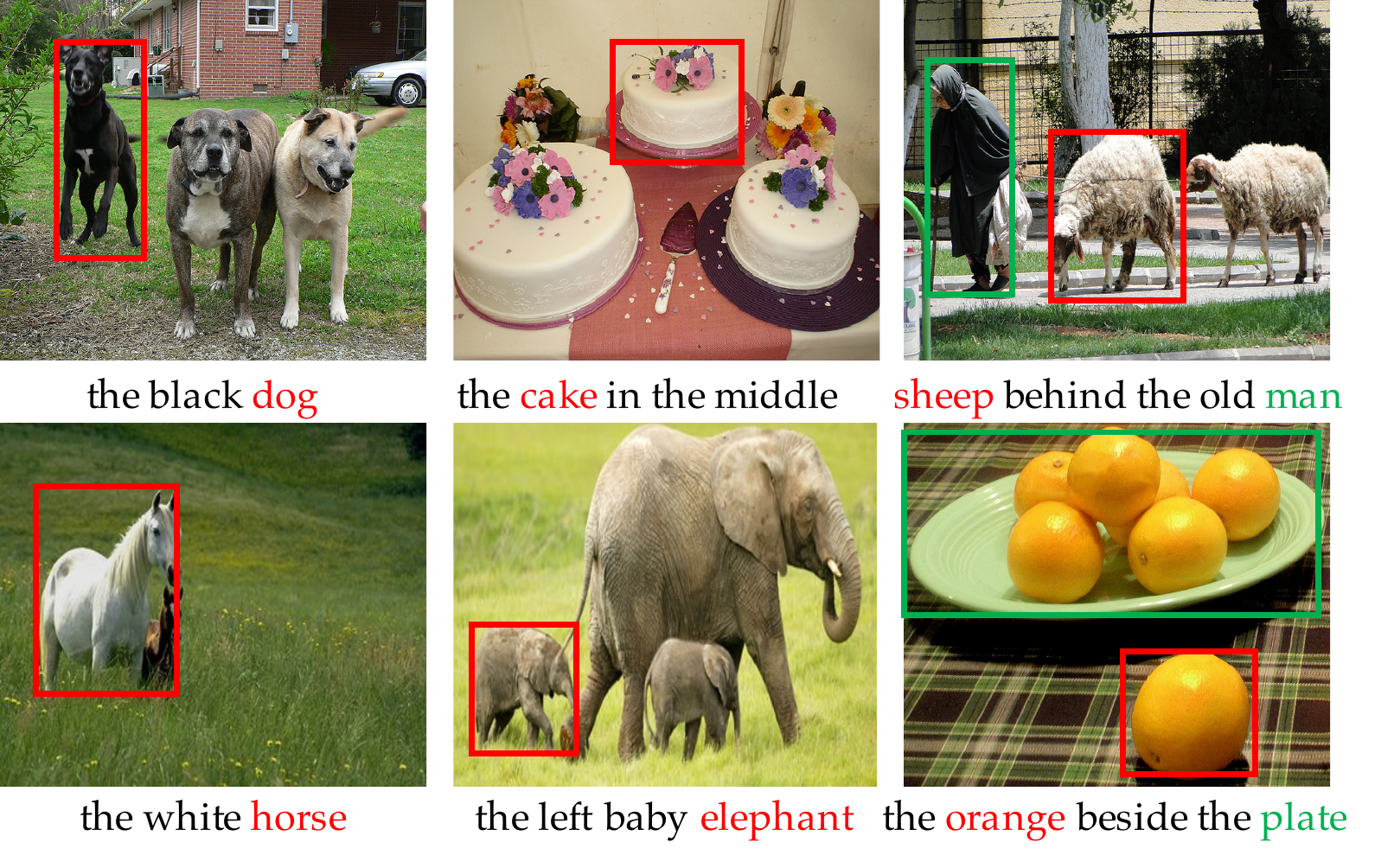}
	\caption{Some examples of referring expression grounding (REG). REG aims to find the particular target in an image described by the language expression. We adaptively learn the subject, location and context because the discriminative information varies in different expressions. 
	Different columns show the dominant cue (subject, location, context) for grounding. 
	 Red box: subject; Green box: object.} 
	\label{fig:examples}
\end{figure}

\IEEEPARstart{W}ith the development of computer vision \cite{DBLP:conf/cvpr/HeZRS16,DBLP:journals/pami/RenHG017,9052469} and natural language processing \cite{DBLP:conf/naacl/DevlinCLT19,DBLP:conf/emnlp/PenningtonSM14}, referring expression grounding (REG) has attracted much attention in recent years \cite{DBLP:conf/cvpr/MaoHTCY016, DBLP:conf/cvpr/Yu0SYLBB18,ACMMM19YSJ,DBLP:journals/tip/WuWSH19,DBLP:conf/cvpr/HuFSZL20,DBLP:conf/eccv/YangC0L20,DBLP:conf/eccv/HuiLHLYZH20}. 
As Fig.~\ref{fig:examples} shows, given a language query (referring expression), REG aims to find the corresponding entity in an image. REG is basic in understanding vision and language, which has wide use in the interaction between humans and computers, such as robotic navigation \cite{DBLP:conf/acl/ThomasonSM17, DBLP:conf/cvpr/AndersonWTB0S0G18,DBLP:conf/eccv/WangJIWKR20}, visual Q\&A \cite{DBLP:journals/pami/LiangJCKLH19,DBLP:journals/pami/LiuXHYS20}, or photo editing \cite{DBLP:journals/tog/ChengZLVSCMT14}.

\begin{figure*}
	\centering
	\includegraphics[width=\textwidth]{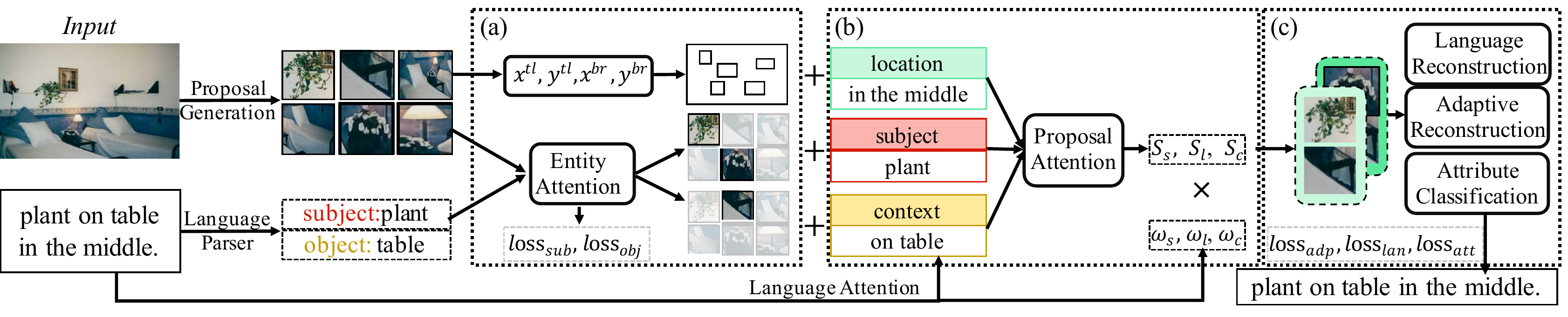}
	\caption{The pipeline of our method. (a) Entity enhancement for proposal selection. (b) Adaptive grounding. (c) Collaborative reconstruction.}
	\label{fig:dataflow}
\end{figure*}
However, training the REG model in a supervised manner requires expensive annotated data to draw the mapping between the queries and proposals. {Limited by the time-consuming process of developing region-level ground-truth, most current REG datasets only contain objects of around 100 categories.} The limited training data hinders the application of REG in the real world. {Weakly supervised REG methods \cite{DBLP:conf/eccv/RohrbachRHDS16,DBLP:conf/cvpr/ChenGN18,DBLP:conf/cvpr/XiaoSL17,DBLP:conf/cvpr/0006LZF18,DBLP:conf/cvpr/ZhangNC18, DBLP:conf/eccv/GuptaVC0KH20} can ground a referential entity with only image-query pairs in training compared with the fully-supervised REG methods.} The image-level annotations are much easier to acquire with Internet search engines.
Hence we focus on weakly supervised REG in this work.

Unlike supervised REG methods, the lack of region-level annotations in weakly supervised REG makes it harder to obtain the correspondence between the query and candidate proposals.
Besides, the key of weakly supervised REG lies in distinguishing a particular object from other objects, usually of the same category. {Most of the existing weakly supervised REG methods only focus on the visual appearance features of the proposals.} The absence of the discriminative location and context information limits their ability to distinguish a particular object from other same-category objects. 
Thus, to implement a weakly supervised REG framework, there exist two main challenges: the first challenge is to introduce meaningful supervision to guide the selection of candidate proposals; the second challenge lies in integrally exploiting the visual appearance, location and context features to distinguish the referent from other objects. 
Further, Yu {\it et al.}~\cite{DBLP:conf/cvpr/Yu0SYLBB18} found that different syntax structures express the variance of referring expression when people refer to an object. 
{Therefore, based on different queries, the grounding system should adaptively focus on the visual appearance, location or context features.}

In light of these observations, we propose an Entity-enhanced Adaptive Reconstruction Network (EARN), a two-stage weakly supervised REG method.
In the first stage, candidate proposals are generated with unsupervised object proposal methods \cite{DBLP:conf/eccv/ZitnickD14} or pre-trained object detection networks \cite{DBLP:conf/nips/RenHGS15,DBLP:conf/cvpr/RedmonF17}. 
In the second stage, we learn the matching scores between the candidate proposals and the linguistic query.  We focus on the second stage here, so this problem can be formulated as follows.
Given an image $I$, a query $q$ and a set of region proposals $\{r_i\}^N_{i=1}$, we need to select the best-matched region $r^*$ according to the query without knowing any $(q, r_i)$ pair. 
{Fig.~\ref{fig:dataflow} shows the network pipeline.}
EARN mainly includes three modules: entity enhancement for proposal selection, adaptive grounding, and collaborative reconstruction.
After acquiring a sequence of candidate proposals, EARN first selects the candidate proposals corresponding to the subject (the target entity) or object (the related entity) under entity enhancement. Then the model learns the mapping between the candidate proposals and queries upon the subject, location and context information in an adaptive manner. 
{The subject covers the visual appearance information, the location handles the absolute and relative location information, and the context learns the relationship between subject and object.}

\textbf{Entity Enhancement for Proposal Selection.} 
This module can select more accurate candidate proposals based on entity enhancement.
First, we extract the subject and object from the referring expression with an external language parser~\cite{DBLP:conf/emnlp/KazemzadehOMB14}.
Then we can predict the category for each candidate proposal with a pre-trained classification network. We calculate the semantic similarity between the word embeddings of subject/object and proposal category in the same space as supervision. 
Under such supervision, we design entity attention to learn the proposal ranking for the subject or object. 
So the model can select the candidate proposals with higher scores and discard the unrelated candidate proposals.

\textbf{Adaptive Grounding.} 
When people refer to a particular target, the language query expresses the variance based on the environment.
For example, ``green plant'', ``the middle plant'' and ``plant on the table'' concentrate on the visual appearance, location and context, respectively.
{To better alleviate the variance of different referring expressions, we adaptively learn the mapping between the query and the proposals of the subject, location and context. }
{Specifically, we put forward a hierarchical attention mechanism, where the first attention calculates ranking scores for the selected candidate proposals of the subject, location and context, and the second attention adaptively assigns different weights to the three different ranking scores to obtain the final matching score between the query and the candidate proposals. }
Besides, we introduce a soft context pooling to encode visual context features, which models the context based on the attentive features of all candidate proposals.
This soft context pooling is more thorough and robust to model context for weakly supervised REG. 

\textbf{Collaborative Reconstruction.} 
Due to the lack of region-level ground-truth, we construct collaborative reconstruction to measure the grounding results by calculating the correspondence between the query and the proposals. 
The collaborative reconstruction consists of the following three parts.
\textit{Adaptive reconstruction} reconstructs attentive hidden features of the subject, location, and context to learn the query-proposal mapping adaptively.
\textit{Language reconstruction} directly reconstructs the input query based on the attentive proposal features, ensuring our model does not neglect any language information.
\textit{Attribute classification} leverages the attribute information of candidate proposal upon the subject to discriminate different proposals.

In summary, the main contribution of this paper is five-fold:
\begin{itemize}
	\item This paper proposes an entity-enhanced adaptive reconstruction network (EARN) for weakly supervised REG without human efforts to collect region-level ground truth for training data.
	
	\item We select the candidate proposals corresponding to subject or object based on entity enhancement, which is based on the semantic similarity as supervision.
	
	\item We propose adaptive grounding with hierarchical attention to adaptively ground the candidate proposals for better handling the variance of different referring expressions.
	
	\item We introduce the collaborative reconstruction to measure the grounding results from three perspectives: adaptive reconstruction, language reconstruction, and attribute classification. 
	
	\item {We conduct experiments to verify different modules and explore different feature encoding and training strategies.	All code and data for EARN is available online~\cite{github_earn}.} 
	
\end{itemize}

EARN extends the previous work ARN~\cite{DBLP:conf/iccv/Liu0WZMH19} and KPRN~\cite{ACMMM19LXJ} in the following aspects: 
 (1) EARN aims to simultaneously address the lack of region-level ground truth and the discrimination of same-category objects in weakly supervised referring expression grounding. 
 (2) We enhance the proposals for subject or object based on semantic similarity. To better model complex relationships, such as multi-entity relationships, we propose soft-context pooling that models the context based on every visual entity in the scene. The comparison is shown in Table~\ref{table:comparison_on_cxtp}. The entity-enhanced soft context pooling is more thorough and robust to model context for weakly supervised REG.
 (3) As shown in Table~\ref{table:coco_acc} and Table~\ref{table:clef_loss}, our method improves the grounding performance on four REG datasets by a large margin, achieving state-of-the-art results on all the datasets. 
 Further, we conduct experiments on a new larger dataset to verify the generalization of our method. 
 (4) We add more ablation experiments to validate the effects of different modules. We also conduct experiments to compare the performance of parsing-based and attention-learned language features, end-to-end training and two-stage method, absolute and relative location features, and different context pooling methods.
{ (5) We enrich the qualitative experiments to show more intuitive visual results. We also show the failure cases and discuss them.}

\begin{figure*}
	\centering
	\includegraphics[width=0.85\textwidth]{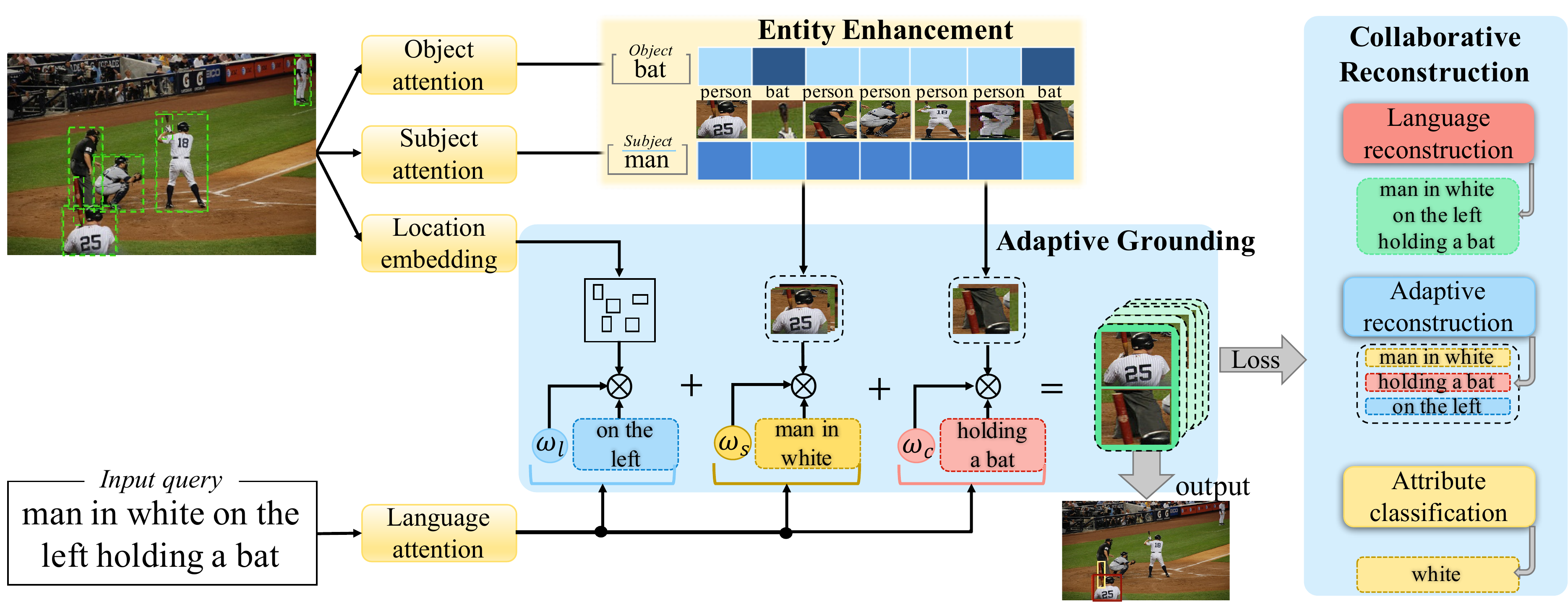}
	\caption{The proposed entity-enhanced adaptive reconstruction network (EARN). Given a query and an image with region proposals, EARN localizes the referential object through adaptive grounding and collaborative reconstruction.}
	\label{model1}
\end{figure*}

\section{Related Work}
\subsection{Supervised Referring Expression Grounding}
REG is also known as referring expression comprehension or phrase localization, which is the inverse task of referring expression generation.
After the release of standard datasets \cite{DBLP:conf/emnlp/KazemzadehOMB14, DBLP:conf/eccv/YuPYBB16, DBLP:conf/cvpr/MaoHTCY016, DBLP:journals/ijcv/PlummerWCCHL17}, REG has made great progress.
Most supervised REG methods can be divided into two kinds.

One is CNN-LSTM based encoder-decoder structure to model $P(q|I,r)$~\cite{DBLP:conf/cvpr/MaoHTCY016, DBLP:conf/eccv/YuPYBB16, DBLP:conf/eccv/NagarajaMD16, DBLP:conf/cvpr/HuXRFSD16, DBLP:conf/cvpr/LuoS17,DBLP:journals/ieeemm/LiJZWH13,DBLP:journals/tip/SongWHT17}.
This kind of method takes extracted CNN features as input to generate language expressions. They aim to maximize the likelihood of the expression for a given referred region at each time step.
Mao {\it et al.}~\cite{DBLP:conf/cvpr/MaoHTCY016} first proposed Maximum Mutual Information (MMI) to train the CNN-LSTM framework. The model would be punished if other objects in the same image could generate the expression.
Besides CNN features, Yu {\it et al.}~\cite{DBLP:conf/eccv/YuPYBB16} introduced visual comparison with other objects in an image to improve the performance.
Nagaraja {\it et al.}~\cite{DBLP:conf/eccv/NagarajaMD16} learned context through multiple instance learning to comprehend referring expressions.
Inspired by the generator-discriminator structure, Luo {\it et al.}~\cite{DBLP:conf/cvpr/LuoS17} utilized referring expression comprehension module as a ''critic`` to generate better expressions.
The CNN-LSTM framework is concise and practical. However, these approaches sequentially encode the expression and ignore the complex dependencies in language expressions.

The other line is the joint vision-language embedding framework to model $P(q, r)$~\cite{DBLP:conf/cvpr/WangLL16, DBLP:conf/iccv/Liu0017, DBLP:conf/iccv/ChenKN17, DBLP:conf/cvpr/Yu0SYLBB18,DBLP:journals/tmm/LiJH12,DBLP:conf/mm/WangCZH018,DBLP:conf/cvpr/HuRADS17}.
This kind of method encodes language and visual features separately. They learn the mapping score between the language and visual features.
Hu {\it et al.}~\cite{DBLP:conf/cvpr/HuRADS17} proposed compositional modular networks to identify entities and relationships to analyze referential expressions.
To avoid ambiguation, Liu {\it et al.}~\cite{DBLP:conf/iccv/Liu0017} incorporated attribute learning into referring expression generation and comprehension to improve the performance.
Yu {\it et al.}~\cite{DBLP:conf/cvpr/Yu0SYLBB18} decomposed the referring expressions into three modules: subject, location and relationship with its environment. 
Based on this work, Liu {\it et al.}~\cite{DBLP:conf/cvpr/LiuWSWL19} designed cross-modal attention-guided erasing to further improve the performance of referring expression grounding. 
This kind of method achieves remarkable results in this direction. The Modular Attention Network (MAttNet) proposed by Yu {\it et al.} has become one of the most popular baselines in REG. 

\subsection{Weakly Supervised Referring Expression Grounding} 

The weakly supervised REG methods can be divided into two kinds, namely single-stage and two-stage methods. The single-stage methods combine the proposal generation and the grounding system. This kind of method can generate more accurate proposals without the dependency on region proposal networks.
Xiao {\it et al.}~\cite{DBLP:conf/cvpr/XiaoSL17} generated attention masks to localize the query by learning the standard discriminative loss and a sentence-based structural loss. 
Zhao {\it et al.}~\cite{DBLP:conf/cvpr/0006LZF18} took region proposals as anchors and learned a multi-scale correspondence network to search for corresponding regions.

Two-stage methods first generate candidate proposals with off-the-shelf region proposal generation networks. 
In the second stage, they calculate the ranking score of the candidate proposals based on the linguistic query. 
Following this pipeline, Rohrbach {\it et al.}~\cite{DBLP:conf/eccv/RohrbachRHDS16} proposed a reconstruction schema that first computes a ranking score of the candidate proposals using attention mechanism, then reconstructs the input query based on the attentive visual features. 
Seeing that this reconstruction method ignored the rich visual information, Chen {\it et al.}~\cite{DBLP:conf/cvpr/ChenGN18} designed knowledge aided consistency network, which reconstructs both the input query and proposal's information.
{Reconstruction is widely used in weakly supervised problems. Duan {\it et al.}~\cite{DBLP:conf/nips/DuanHGW0H18} solved the weakly supervised dense event caption problem by decomposing it into dual problems: event captioning and sentence localization.}
Different from the above reconstruction-based methods, Zhang {\it et al.}~\cite{DBLP:conf/nips/ZhangZLZH20} creatively put forward counterfactual contrastive learning (CCL) to develop sufficient contrastive training between counterfactual positive and negative results. 
These methods can dig deeper into the relationship between the candidate proposals.
EARN belongs to the two-stage method, which aims to calculate the ranking score of generated candidate proposals.
\section{Method}
We propose an entity-enhanced adaptive reconstruction network (EARN) to address the weakly supervised REG problem.
EARN is two-stage method. In the first stage, we acquire a set of candidate proposals through region proposal or detection networks. In the second stage, EARN chooses the most probable proposal from these candidate proposals.
Specifically, EARN adaptive grounds the target proposal under the assistance of introduced supervision, then reconstructs its corresponding query with a collaborative loss.  The whole network architecture is shown in Fig. \ref{model1}.


\subsection{Feature Encoding}
\label{feats_enc}

	This subsection introduces the visual and language features extraction in detail.
	Visual features are extracted from external networks or arithmetic operators for region proposals. 
	The referring expression features are extracted through LSTM and attention mechanism. 

\subsubsection{RoI Features}

\textbf{Subject feature} is defined as the visual representations of proposals.
We crop the C3 and C4 features of Faster R-CNN based on ResNet \cite{DBLP:conf/cvpr/HeZRS16} after runing the forward propagation for each image as the subj ect feature $\widetilde { r } _ { s } ^ { i } = f_{CNN}(r_i)$. The C3 features contain lower-level features like colors and shapes. The C4 features represent more higher-level representations. 

\textbf{Location feature} is the combination of each proposal's absolute and relative location features.
The absolute location feature represents the position of each proposal in the image. The relative location feature is the relative location between each proposal and other proposals of the same category in the image. Following \cite{DBLP:conf/eccv/YuPYBB16, DBLP:conf/cvpr/YuTBB17, DBLP:conf/cvpr/Yu0SYLBB18}, we decode the absolute location feature of each proposal as a 5-dim vector $r_l^i = \left[ \frac { x _ { t l } } { W } , \frac { y _ { t l } } { H } , \frac { x _ { b r } } { W } , \frac { y _ { b r } } { H } , \frac { w \cdot h } { W \cdot H } \right]$, denoting the top-left, bottom-right position and relative area of the proposal to the whole image. 
The relative location feature is the relative location between the proposal and five surrounding proposals of the same category. 
For each surrounding proposal, the offset and area ratio to the candidate is calculated as: $\delta r_l^{ij} = \left[ \frac { \left[ \Delta x _ { t l } \right] _ { i j } } { w _ { i } } , \frac { \left[ \triangle y _ { t l } \right] _ { i j } } { h _ { i } } , \frac { \left[ \triangle x _ { b r } \right] _ { i j } } { w _ { i } } , \frac { \left[ \triangle y _ { b r } \right] _ { i j } } { h _ { i } } , \frac { w _ { j } h _ { j } } { w _ { i } h _ { i } } \right]$.
{Finally, the above absolute and relative location features are concatenated as the location feature of the proposal, which is a 30-dim vector: $\widetilde { r } _ { l } ^ { i } = \left[ r _ { l }^{i} ; \delta r _ { l }^{i} \right] $.}

\textbf{Context feature} is the relationship between the candidate proposal and other objects in the image.
The feature of the surrounding proposal contains C4 feature $v_{ij}=f_{CNN}(r_{j})$ and its relative location feature. The relative location feature is encoded as follows: $\delta m _ { i j } = \left[ \frac { \left[ \triangle x _ { t l } \right] _ { i j } } { w _ { i } } , \frac { \left[ \triangle y _ { t l } \right] _ { i j } } { h _ { i } } , \frac { \left[ \triangle x _ { b r } \right] _ { i j } } { w _ { i } } , \frac { \left[ \triangle y _ { b r } \right] _ { i j } } { h _ { i } } , \frac { w _ { j } h _ { j } } { w _ { i } h _ { i } } \right]$. The context feature is calculated as $\widetilde { r } _ { c } ^ { ij } =  \left[ v _ { i j } ; \delta m _ { i j } \right]$. 


\subsubsection{Referring Expression Features}
\label{reffeats}
{The referring expression features of subject $q_s$, location $q_l$ and context $q_c$ are learned with language attention.	}
Specifically, we first encode each word in a query $q = \left\{ w _ { t } \right\} _ { t = 1 } ^ { T }$ into a one-hot vector and then map it as a word embedding $e_t$. Then, the word embedding $e_t$ is fed into a bi-directional LSTM. The hidden vectors in both directions are concatenated as the final representation $h_t = [\overrightarrow{h}_t, \overleftarrow{h}_t]$. 
We use an attention mechanism to learn the representation of the subject, location, and context.
{Taking subject feature $q_s$ as an example, we calculate its hidden representation as follows:}
\begin{equation}
\begin{aligned}
{ m } _ { t } &= \mathrm { fc } \left(  { h } _ { t } \right) ,\\
\alpha _ { t } &= \operatorname { softmax } _ { t } \left( { m } _ { t } \right) ,\\
{ q_s } &= \sum _ { t } \alpha _ { t }  { e } _ { t }.
\end{aligned}
\label{att_sub}
\end{equation}

The location feature $q_l$ and context feature $q_c$ are calculated in the same way.
Besides, we calculate three different weights upon the subject, location and context with the hidden state vector of the bi-directional LSTM to represent the importance of each part (subject, location and context). The operations are shown in Eq. (\ref{weights}).
\begin{equation}
\left[ w _ { s } , w _ { l  } , w _ { c } \right] = \operatorname { softmax }_{w} \left( \mathrm { fc } \left( \left[{ h } _ { 0 }, { h } _ { T } \right] \right) \right)
\label{weights}
\end{equation}

\subsection{Entity Enhancement for Proposal Selection}
We propose entity enhancement to highlight the candidate proposals for the subject or object under calculated supervision.
The supervision is in the form of semantic similarity between the proposal category and the subject/object to guide the selection of candidate proposals. 

{Specifically, the category $C_i$ for each image proposal is extracted with pre-trained Faster R-CNN~\cite{DBLP:conf/nips/RenHGS15}. }
Then, we parse the referring expression into seven attributes: category name, color, size, absolute location, relative location, relative object, and generic attribute~\cite{DBLP:conf/emnlp/KazemzadehOMB14}. We define the category name as the subject $W_s$, and the relative object as the object $W_o$.
The proposal category ($C_i$) and subject ($W_s$)/object ($W_o$) are encoded into vectors $emb_c$ and  $emb_s$, $emb_o$ according to the GloVe pre-trained word vectors~\cite{DBLP:conf/emnlp/PenningtonSM14}.
``unk''  is used to indicate the word which is out of the vocabulary.
We calculate the cosine distance between the proposal category and subject/object as semantic similarities $SIM_s$ and $SIM_o$ after they are encoded into the same semantic space.

{We use the semantic similarity as supervision for entity attention to calculate initial proposal ranking scores.} The entity attention includes subject and object attention.
{Through subject attention, the candidate proposals that have little probability of being the target are assigned to lower weights or even excluded in the next processor. }
Object attention calculates the weights for the candidate proposal contributing to the context.

Taking the subject as an example, the candidate proposal features and the embedding of the subject are first concatenated into one vector. Then the vector is fed into a two-layer perceptron, which is the subject attention. Finally, we calculate the corresponding matching score between proposals and the subject. Eq. (\ref{eq.sub_attn}) shows the process of subject and object attention. 

\begin{equation}
\begin{aligned}
\overline{Score}_s^{i}&= f_{ATT}\left(\widetilde { r } _ { s } ^ { i } ,emb_s\right) =\space W _ { 2 } \phi_{\rm ReLU} \left( W _ { 1 } [ \widetilde { r } _ { s } ^ { i } ,emb_s] \right),\\
\overline{Score}_o^{{ij}}&= f_{ATT}\left(v_{ij},emb_o\right) =\space W _ { 2 } \phi_{\rm ReLU} \left( W _ { 1 } [ v_{ij},emb_o] \right),
\end{aligned}
\label{eq.sub_attn}
\end{equation}
where $\widetilde { r } _ { s } ^ { i }$, $ v_{ij}$ denote the subject and object features extracted from the proposals in the image; $emb_s$ and $emb_o$ represent the embedding of subject and object from the query. 
Then, the scores are normalized with softmax as $Score_s^i $, $Score_o^{ij}$.

The calculated semantic similarity is the supervision for the subject and object attention networks.
{We utilize the mean squared error (MSE) between the matching score and the semantic similarity as follows:}
\begin{equation}
\begin{aligned}
Loss_{sub} &={\rm MSE}\left(Score_s, SIM_s\right),\\
Loss_{obj} &= {\rm MSE}\left(Score_o, SIM_o\right).
\end{aligned}
\end{equation}

{We use two kinds of methods to select the candidate subject proposals based on the subject attention score. } One is to assign different weights to the candidate proposals when grounding the query, called as \textbf{soft filter}. Specifically,  we multiply the final ranking score $S_t^i $ with the subject attention score $Score_s^i$ in the soft filter. The other is to discard the unrelated proposals if their subject attention score is under the threshold, named as \textbf{hard filter}.

{We consider three different methods to encode context proposal features.} In the first version, we choose five surrounding proposals of the target in different categories as candidate context proposals. From the above candidate proposals, we finally select the one with the maximum response to the query as the final relative object, denoted as $\widetilde { r } _ { c } ^ { i }$.
{The second kind of context pooling regards all the proposals in the image as candidate proposals.} Based on the learned object attention $Score_o^{ij}$, we select the one with maximum score as the context proposal $\widetilde { r } _ { c } ^ { i }$. 
Different from the above two max pooling methods, we put forward soft context pooling as the third way to encode context proposal features. The context is modeled based on the features of all candidate proposals. The weights are the learned object matching scores. The final context embedding is the linear combination of all the candidate proposal features as: 			
\begin{equation}
\widetilde { r } _ { c } ^ { i  } = \sum _ { j = 1 } ^ { N } Score_o^{ij} \widetilde { r } _ { c } ^ { ij }.
\end{equation} 
This context pooling explores all the proposals in the image, which can better handle multi-entity relationships.

\subsection{Adaptive Grounding}
\label{adp_loc}
We propose hierarchical attention to localize a query based on the subject, location and context features. 
The first \textbf{proposal attention} calculates the matching score between the proposals and query upon subject, location, and context respectively. 
The second \textbf{language attention} assigns different weights to subject, location and context based on the query to alleviate variance in queries.

\begin{figure}[]
	\centering
	\subfigure[Adaptive Grounding.]{
		\label{grounding}
		\includegraphics[height=5cm]{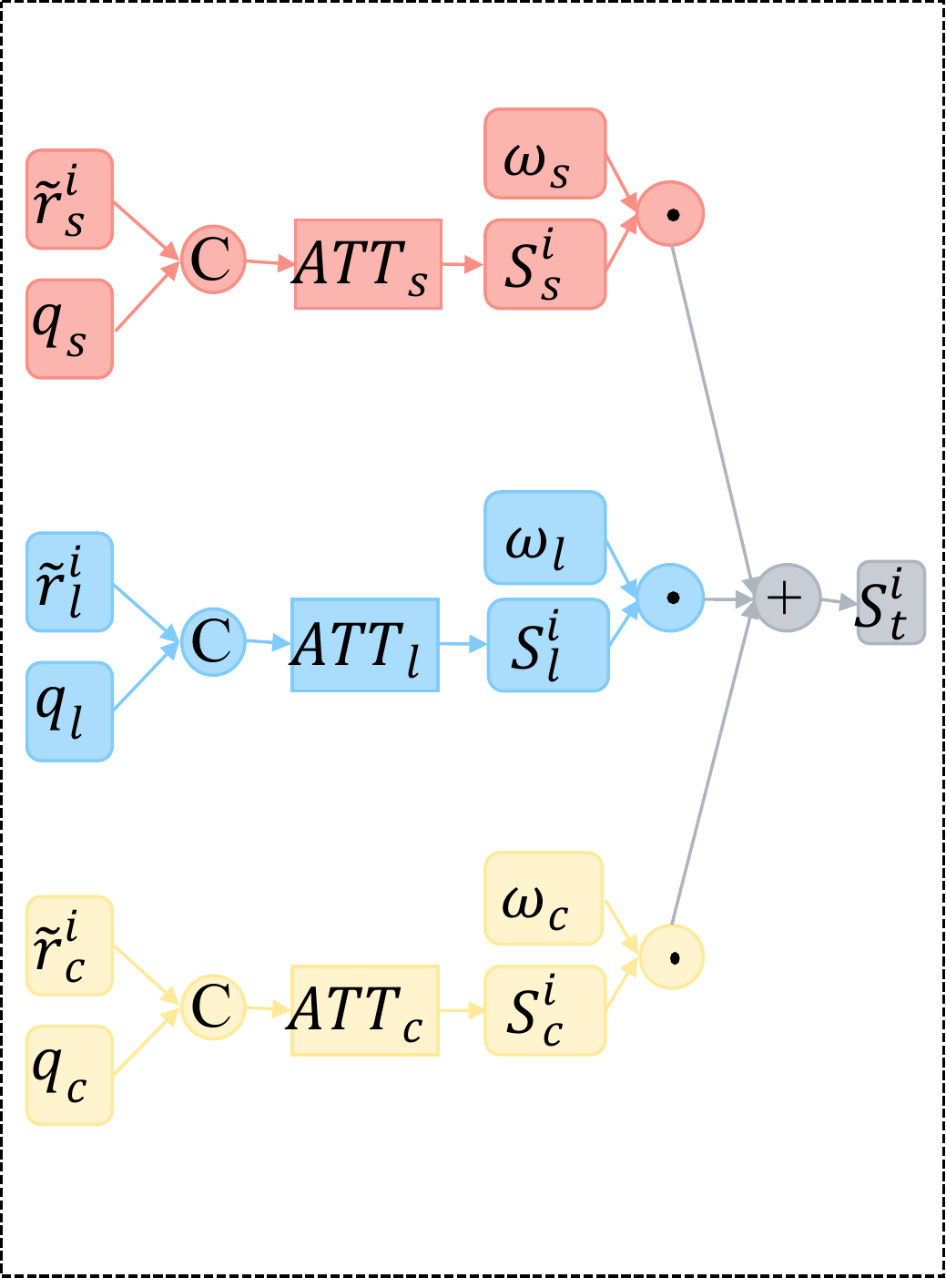}}
	\subfigure[Collaborative Reconstruction.]{
		\label{reconstruct}
		\includegraphics[height=5cm]{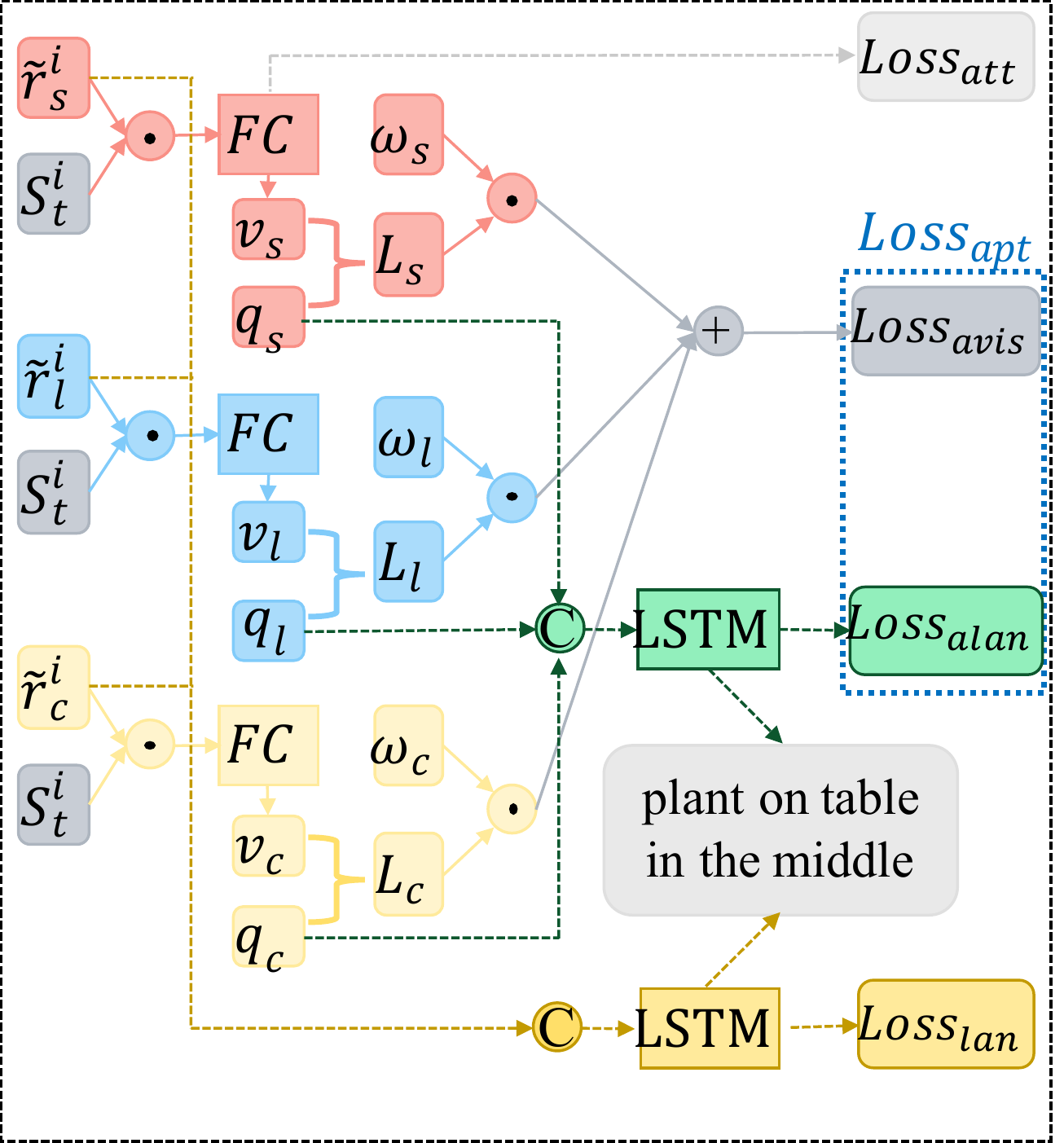}}
	\caption{The structure of adaptive grounding (Section \ref{adp_loc}) and collaborative reconstruction (Section \ref{adp_rec}). The reconstruction process contains attribute classification loss, adaptive reconstruction loss (including adaptive language and visual reconstruction loss) and language reconstruction loss. \textit{ATT}: attention layer. $\oplus$: plus operation. $\odot$: element-wise vector multiplication. C: vector concatenation.}
	\label{model2}
	\vspace{1.2em}
\end{figure}

Fig. \ref{grounding} shows the grounding process. $\widetilde { r } _ { s } ^ { i }$, $\widetilde { r } _ { l } ^ { i }$ and $\widetilde { r } _ { c } ^ { i }$ denote the visual features extracted from the region proposals in the image through CNN. $q_s$, $q_l$ and $q_c$ are the language features extracted from the query through bi-directional LSTM.
Taking the subject as an example, $\widetilde { r } _ { s } ^ { i }$ and $q_s$ are first concatenated into one vector. Then the vector is fed into the proposal attention, which is a two layer perceptron, to learn the corresponding matching score. The whole pipeline is formulated in Eq. (\ref{subscore}), where the biases are omitted.
\begin{equation}
\label{subscore}
{  \overline{ s } } _ { x }^{i} = f _ { A T T } \left( q_x , \widetilde { r } _ { x } ^ { i } \right) = W _ { 2 } \phi_{\rm ReLU} \left( W _ { 1 } [ q_x , \widetilde { r } _ { x } ^ { i }] \right), x \in (s, l, c)
\end{equation}

We then normalize the scores with softmax as:
\begin{equation}
{ s } _ { x } ^ { i }  = \operatorname { softmax } _ { i } \left( \overline { s } _ { x } ^ { i } \right), \quad x \in (s, l, c). 
\end{equation}
Based on language attention, the final ranking score of candidate proposals is the linear combination of the three sub-score:
\begin{equation}
S_t^i = w _ { s }s _ { s } ^ { i } + w _ { l } s _ { l } ^ { i } + w _ { c } s _ { c } ^ { i }.
\end{equation}
The final score represents the probability of region \textit{i} matching query \textit{q} considering subject, location and context.
The weights show the descriptive tendency of the referring expression.

\subsection{Collaborative Reconstruction}
\label{adp_rec}

{Weakly supervised REG lacks the region-level ground-truth, so we construct a collaborative reconstruction in the training stage to measure the adaptive grounding results.} 
The collaborative loss includes three losses, as shown in Fig.~\ref{reconstruct}.
The adaptive reconstruction reconstructs attentive hidden features of subject, location and context respectively. 
The language reconstruction directly reconstructs the input query based on the attentive features of proposals.  
The attribute classification takes advantage of attribute information of the referential proposal.

\subsubsection{Adaptive Reconstruction Loss}
We build adaptive reconstruction to bridge the semantic gap between input query and proposal. As Fig. \ref{adaploss} shows, the adaptive grounding reconstructs different linguistic queries with corresponding proposal features to better handle the variance among different expressions in the datasets. 
This loss consists of two sub-losses, adaptive visual reconstruction loss and adaptive language reconstruction loss.
The first loss helps to reconstruct query features $q _ { s }$, $q _ { l }$ and $q _ { c }$ with proposal features $\widetilde { r } _ { s } ^ { i }$, $\widetilde { r } _ { l } ^ { i }$ and $\widetilde { r } _ { c } ^ { i }$. 
We first aggregate the visual features based on their corresponding proposal ranking score to learn attentive proposal features as follows:
\begin{figure}
	\centering
	\subfigure[]{
		\label{adaploss}
		\includegraphics[height=5cm]{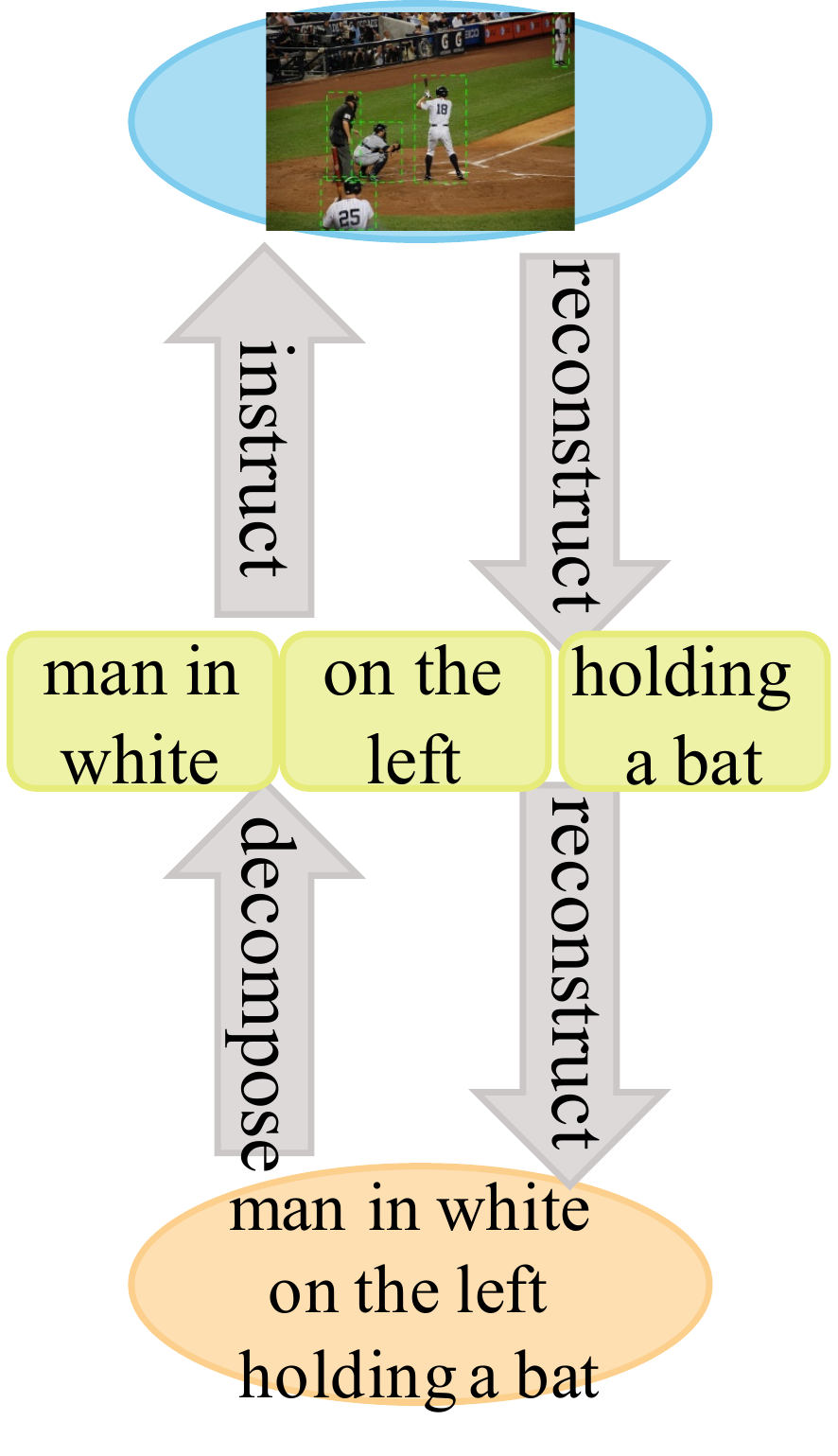}
	}
	\hspace{0.2 cm}
	\subfigure[]{
		\label{langloss}
		\includegraphics[height=5cm]{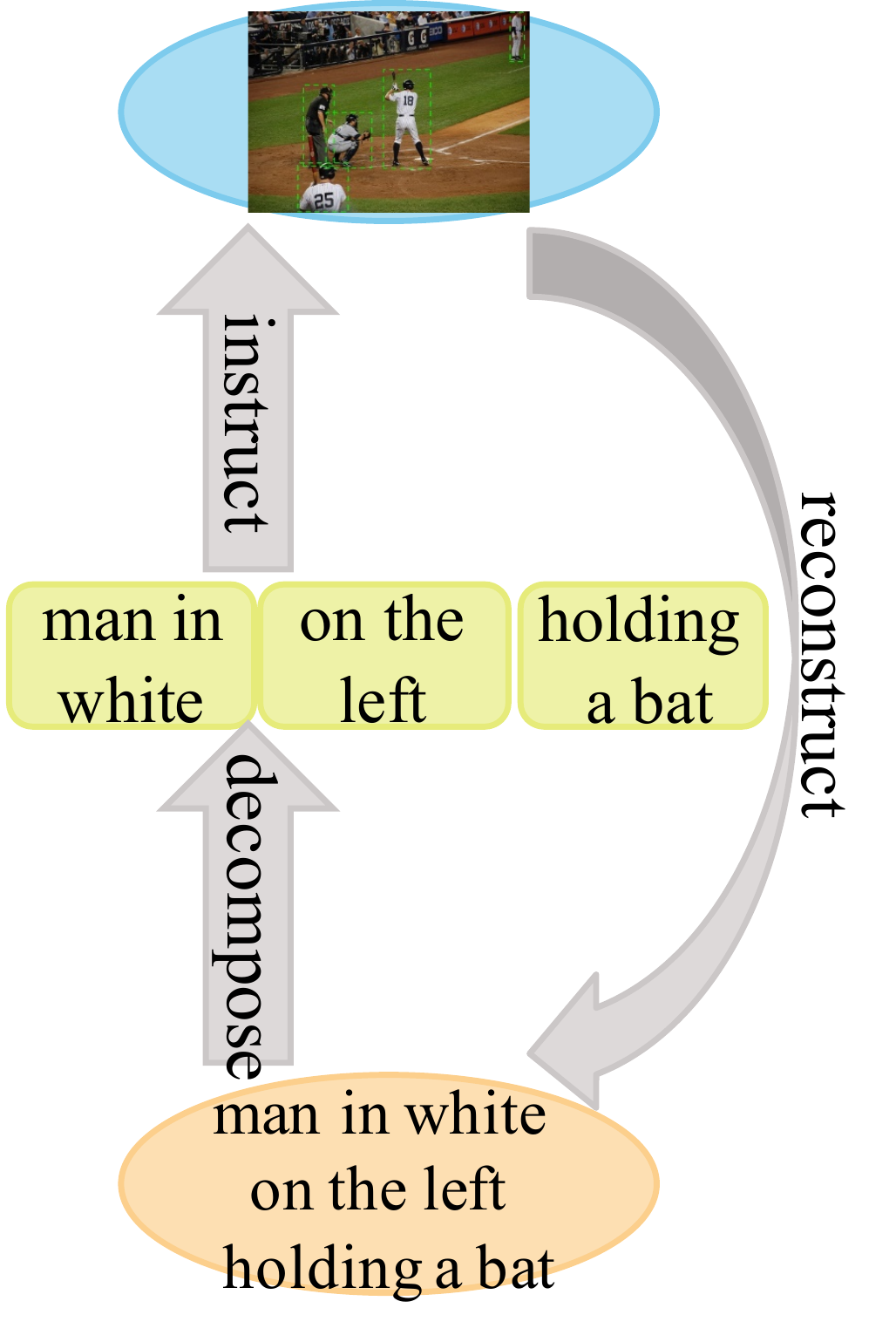}}
	\caption{The sketch map of (a) Adaptive reconstruction and (b) Language reconstruction.}
	\label{RecLoss}
\end{figure}

\begin{equation}
\tilde {v} _ { s } = \sum _ { i = 1 } ^ { N } S _ { t } ^ { i } \tilde { r } _ { s } ^ { i },\quad
\tilde {v} _ { l } = \sum _ { i = 1 } ^ { N } S _ { t } ^ { i } \tilde { r } _ { l } ^ { i },\quad
\tilde {v} _ { c } = \sum _ { i = 1 } ^ { N } S _ { t } ^ { i } \tilde { r } _ { c } ^ { i }.
\end{equation}
Then the attentive proposal features are fed into a fully connected layer to be embedded in common semantic space with the language features:
\begin{equation}
v _ { s } = FC(\tilde {v} _ { s }),\quad
v _ { l } = FC(\tilde {v} _ { l }),\quad
v _ { c } = FC(\tilde {v} _ { c }).
\end{equation}
Then we use the attentive proposal features $v_s$, $v_l$ and $v_c$ to reconstruct the language features $q _ { s }$, $q _ { l }$ and $q _ { c }$ extracted from the original query. {We use the MSE between the distance between the proposal features and language features for the loss term:}
\begin{equation}
L_x ={\rm MSE}(v _ { x }, q_x), \quad x \in (s, l, c).
\label{eq.mse}
\end{equation}

The final adaptive visual reconstruction loss is the weighted sum of the subject reconstruction loss, location reconstruction loss and context reconstruction loss as:
\begin{equation}
Loss_{avis} = w _ { s } L_s + w _ { l } L_l + w _ { c } L_c, 
\end{equation}
where the weights are calculated based on the query, as Eq.(\ref{weights}) shows.

The language feature extraction network is trained together with the grounding and reconstruction network. $v _ { x }$ are the attentive proposal features, and $q_x$ are the learned language features. As $v _ { x }$ and $q_x$ are jointly learned by the network, the network parameters might be set to zero roughly to reach convergence as soon as possible. If so, the network can not learn the correspondence between the visual and language modality. 
To avoid this problem, we add an adaptive language reconstruction module to reconstruct the original query with the language features $q _ { s }$, $q _ { l }$ and $q _ { c }$. 
First we concatenate $q _ { s }$ , $q _ { l }$ and $q _ { c }$, then feed it into a one-layer perceptron:
\begin{equation}
f_{alan}=\phi _ {\rm R e L U } \left(W_l( \left[ q _ { s } , q_l, q_c \right])+b_l  \right).
\end{equation}
{Inspired by the query generation methods~\cite{DBLP:conf/cvpr/DonahueHGRVDS15,DBLP:conf/cvpr/VinyalsTBE15}, we reconstruct the input query through LSTM based on the fused language features $f_{lan}$.} As shown in Eq. (\ref{eq.alan}) , the language features $f_{lan}$ are fed into a one-layer LSTM only at the first time step.
\begin{equation}
P ( q | f_{alan} ) = f _ {\rm L S T M } \left(f_{alan} \right)
\label{eq.alan}
\end{equation}
The language reconstruction maximizes the likelihood of the ground-truth query $\hat { q }$ generated by LSTM as:
\begin{equation}
\label{lanloss}
Loss _ { alan } = - \frac { 1 } { B } \sum _ { b = 1 } ^ { B } \log ( P ( \hat { q } | f_{alan} ) ) ,
\end{equation}
where B denotes the batch size.
The final adaptive reconstruction loss is the linear combination of the language reconstruction loss and visual reconstruction loss as Eq. (\ref{eq.adp}) shows. 
\begin{equation}
Loss_{adp} = \alpha Loss_{avis}  + \beta  Loss_{alan},
\label{eq.adp}
\end{equation}
where $\alpha$ and $\beta$ denote the hyper-parameters representing the proportion of the two losses. 
\subsubsection{Language Reconstruction Loss}
The language reconstruction module directly reconstructs the input query based on the attentive proposal features as Fig. \ref{langloss} shows. First, the concatenation of the original proposal features  $\widetilde { r } _ { s } ^ { i }$, $\widetilde { r } _ { l } ^ { i }$ and $\widetilde { r } _ { c } ^ { i }$ are fed into a one-layer perceptron as follows:
\begin{equation}
r_{vis}^i=\phi _ {\rm R e L U } \left(W_v( \left[ \widetilde { r } _ { s } ^ { i }, \widetilde { r } _ { l } ^ { i }, \widetilde { r } _ { c } ^ { i } \right])+b_v  \right).
\end{equation}
Then we calculate the weighted sum of the proposal features based on the total score as:
\begin{equation}
{ f } _ { vis } = \sum _ { i = 1 } ^ { N } S _ { t } ^ { i }  { r } _ { vis } ^ { i }.
\end{equation}
Based on the fused proposal features, queries are generated using LSTM:
\begin{equation}
P ( q | f_{vis} ) = f _ {\rm L S T M } \left(f_{vis} \right).
\end{equation}
We use the same language reconstruction loss as Eq. (\ref{lanloss}) in the following step:
\begin{equation}
Loss _ { lan } = - \frac { 1 } { B } \sum _ { b = 1 } ^ { B } \log ( P ( \hat { q } | f_{vis} ) ).
\label{eq.lan}
\end{equation}
Compared with adaptive reconstruction, the language reconstruction reconstructs the input query directly, so it will not damage any useful language information during training.
\subsubsection{Attribute Classification Loss}
As mentioned in previous methods \cite{DBLP:conf/iccv/YaoPLQM17,DBLP:journals/pami/WuSWDH18, DBLP:conf/cvpr/Yu0SYLBB18}, attribute information is essential to distinguish the object of the same category. Here, we introduce an attribute classification branch in our model. 
The attribute labels are extracted with an external language parser \cite{DBLP:conf/emnlp/KazemzadehOMB14} according to \cite{DBLP:conf/cvpr/Yu0SYLBB18}.
Subject feature $\widetilde { r } _ { s } ^ { i }$ of proposal is used for attribute classification.
As each query has multiple attribute labels, we use the binary cross-entropy loss for the multi-label classification:
\begin{equation}
Loss _ { a t t  } = f_{\rm BCE} \left( y _ { i j }, p_{ij} \right),
\end{equation}
where $y_{ij}$ indicates the attribute labels extracted with external language parser; $p_{ij}$ means the predicted attributes; $i$ denotes the i-th example, and $j$ represents the j-th attribute. 
We use the reciprocal frequency that attribute labels appear as weights in this loss to ease unbalanced data.


\subsection{Training and Inference}
The referring expression feature extraction, the subject and object attention, the adaptive grounding and the collaborative reconstruction module are jointly trained in the training stage. Only queries with attribute words are fed into the attribute classification branch during training. 
The collaborative reconstruction loss is defined as:
\begin{equation}
Loss_{clb} = Loss_{adp} + \gamma Loss_{lan} +\lambda Loss_{att}
\end{equation}
The final loss function is formulated as:
\begin{equation}
Loss = Loss_{sub} +  Loss_{obj} + Loss_{clb}
\end{equation}
At inference, the reconstruction module is not needed anymore. We feed the image and query into the network, and acquire the most related proposal whose final score is the maximal in the grounding module as Eq. (\ref{eq.max}) shows.
\begin{equation}
j = \arg \max _ { i } f  \left( p , r _ { i } \right)
\label{eq.max}
\end{equation}

\section{Experiment}
	\begin{table*}[]
	\centering
		\centering
		\caption{{Accuracy (IoU $>$ 0.5) of comparing methods on RefCOCO, RefCOCO+ and RefCOCOg datasets. \textbf{Bold}: best result.  'det' denotes the candidate proposals are generated by detection network, otherwise from the dataset.}}
		\begin{tabular}{c|c|ccc|ccc|c}
			\hline
			\multirow{2}{*}{Methods}&\multirow{2}{*}{Settings}&\multicolumn{3}{c|}{RefCOCO}&\multicolumn{3}{c|}{RefCOCO+}& RefCOCOg\\ 
			\cline{3-9}&&val&testA&testB&val&testA&testB& val\\ \hline 
			VC (det)~\cite{DBLP:conf/cvpr/ZhangNC18} 
			& w/o reg & - & 17.14 & 22.30 & - & 19.74 & 24.05 & 28.14 \\ 
			VC (det)&- & - & 20.91 & 21.77 & - & 25.79 & 25.54 & {33.66} \\ 
			VC (det)& w/o $\alpha$  & - & {32.68} & {27.22} & - & {34.68} & {28.10} & 29.65 \\ \hline \hline
			VC & w/o reg& - &13.59& 21.65& - & 18.79 & 24.14 & 25.14 \\ 
			VC& -& - & 17.34 & 20.98 & - & 23.24 & 24.91 & {33.79} \\ 
			VC & w/o $\alpha$ & - & {33.29} & {30.13} & - & {34.60} & {31.58} & 30.26 \\ \hline
			IGN~\cite{DBLP:conf/nips/ZhangZLZH20} 
			 &Base&31.05&34.39&28.16&31.13&34.44&29.59&32.17 \\
			IGN  &CCL&{34.78}&{37.64}&{32.59}&{34.29}&{36.91}&{33.56}&{34.92} \\ \hline
			ARN (det)~\cite{DBLP:conf/iccv/Liu0WZMH19}
			 & $L_{lan}+L_{adp}+L_{att}$  &32.17&35.35&30.28&32.78&34.35&32.13&33.09\\
			ARN
			 & $L_{lan}+L_{adp}+L_{att}$  &{34.26}&{36.01}&{33.07}&{34.53}&{36.01}&{33.75}&{34.66} \\
			KPRN~\cite{DBLP:conf/mm/LiuLWZSH19} 
			& $L_{lan}+L_{adp}+L_{att}$  &{36.34}&{35.28}&{37.72}&{37.16}&{36.06}&\textbf{{39.29}}&{38.37} \\ \hline
			EARN& $L_{adp}$ & 35.31& 37.07& 32.66 & 35.50 & 37.39& 33.65&38.99\\
			EARN& w/o $L_{adp}$ & 34.93& 33.76& 36.98 & 35.31 & 33.46& 37.27&38.37 \\
			\multirow{1}{*}{EARN} &  
			\multirow{1}{*}{$L_{lan}+L_{adp}+L_{att}$ }
			& {\textbf{38.08}}& {\textbf{38.25}} & { \textbf{38.59}} & \textbf{37.54} & {\textbf{37.58 }}& {37.92 }& {\textbf{45.33}} \\ 
			\hline	
		\end{tabular}
		\label{table:coco_acc}
\end{table*}
\begin{table}[]
	\centering
	\caption{Accuracy (IoU $>$ 0.5) of comparing methods on RefCLEF and Flickr30k dataset.}
	\scalebox{1.05}{	\begin{tabular}{l|c|c}
			\hline
			Method  & RefCLEF&Flickr30k  \\ \hline
			LRCN \cite{DBLP:conf/cvpr/DonahueHGRVDS15}   & 8.59 &-    \\ 
			Caffe-7K \cite{DBLP:conf/rss/GuadarramaRSZFD14} & 10.38 & -  \\ 
			GroundeR \cite{DBLP:conf/eccv/RohrbachRHDS16} & 28.94   \\ 
			MATN  \cite{DBLP:conf/cvpr/0006LZF18} & 13.61 &33.10  \\ 
			VC \cite{DBLP:conf/cvpr/ZhangNC18} & 14.11&-\\
			VC w/o $\alpha$ \cite{DBLP:conf/cvpr/ZhangNC18} & 14.50&-\\
			KAC Net \cite{DBLP:conf/cvpr/ChenGN18} & 15.83 &38.71  \\ \hline \hline
			ARN~\cite{DBLP:conf/iccv/Liu0WZMH19}  & 26.19&37.95 \\
			EARN  &36.86 &38.73 \\ \hline
	\end{tabular}}
	\label{table:clef_acc}
\end{table}

\subsection{Datasets}
We evaluate our method on five popular benchmarks of referring expression grounding.


\noindent \textbf{RefCOCO \cite{DBLP:conf/eccv/YuPYBB16}.} 
The dataset contains 142,209 queries for 50,000 objects in 19,994 images from MSCOCO \cite{DBLP:conf/eccv/LinMBHPRDZ14}.
The dataset is split into train, validation, Test A, and Test B, with 16,994, 1,500, 750 and 750 images, respectively. Test A contains multiple people while Test B contains multiple objects. Each image contains at least 2 objects of the same object category. 

\noindent \textbf{RefCOCO+ \cite{DBLP:conf/eccv/YuPYBB16}.} 
It has 141,564 queries for 49,856 referents in 19,992 images from MSCOCO \cite{DBLP:conf/eccv/LinMBHPRDZ14}. 
Unlike RefCOCO, the queries in this dataset are disallowed to use locations to describe the referents. 
The split is 16,992, 1,500, 750 and 750 images for train, validation, Test A, and Test B respectively. Each image contains 2 or more objects of the same object category in this dataset. 

\noindent \textbf{RefCOCOg \cite{DBLP:conf/cvpr/MaoHTCY016}.} 
It has 95,010 queries for 49,822 objects in 25,799 images from MSCOCO \cite{DBLP:conf/eccv/LinMBHPRDZ14}. 
It has longer queries containing appearance and location to describe the referents. 
The split is 21,149 and 4,650 images for training and validation. There is no open test split for RefCOCOg. Images were selected to contain between 2 and 4 objects of the same category. 

\noindent \textbf{RefCLEF \cite{DBLP:conf/emnlp/KazemzadehOMB14}.} 
It contains 20,000 annotated images from IAPR TC-12 dataset \cite{article} and SAIAPR-12 dataset \cite{DBLP:journals/cviu/EscalanteHGLMMSPG10}. The dataset includes some ambiguous queries, such as anywhere. It also has some mistakenly annotated image regions. The dataset is split into 9,000, 1,000 and 10,000 images for training, validation and test for fair comparison with \cite{DBLP:conf/eccv/RohrbachRHDS16}. 100 bounding box proposals \cite{DBLP:conf/cvpr/HuXRFSD16} are provided for each image using Edge Boxes \cite{DBLP:conf/eccv/ZitnickD14}. Images contain between 2 and 4 objects of the same object category. The maximum length of all the queries is 19 words.

\noindent \textbf{Flickr30K Entities \cite{DBLP:journals/ijcv/PlummerWCCHL17}.} 
This dataset contains 31,783 images with 276K bounding boxes. Each image is associated with 5 descriptive sentences.  Different from the above datasets, the queries in Flickr30k Entities mainly include the category of the referred object. Besides, it does not have the constraint that an image should contain multiple same-category objects. Thus context may be unnecessary to distinguish the target object for this dataset.

\subsection{Experimental Setup}
\subsubsection{Implementation Details}
{EARN is trained through Adam \cite{DBLP:journals/corr/KingmaB14} algorithm with an initial learning rate 4e-4, which is dropped by 10 after every 8,000 iterations.} The training iterations are up to 30,000 with a batch size of a single image. Each image has an indefinite number of annotated queries. 
ResNet is our main feature extractor for RoI visual features. 
We adopt EdgeBoxes \cite{DBLP:conf/eccv/ZitnickD14} to generate 100 region proposals for RefCLEF dataset for fair comparison with \cite{DBLP:conf/eccv/RohrbachRHDS16, DBLP:conf/cvpr/ChenGN18}. 


\subsubsection{Metrics}
The Intersection over Union (IoU) between the selected region and the ground-truth is calculated to evaluate the network performance. Same with previous works, if the IoU score is greater than 0.5, the predicted region is considered as the right grounding.

\subsection{Comparison with the State-of-the-Art}
We evaluate our method in two aspects. First, we compare our method with state-of-the-art weakly supervised methods, including LRCN \cite{DBLP:conf/cvpr/DonahueHGRVDS15}, Caffe-7K \cite{DBLP:conf/rss/GuadarramaRSZFD14}, GroundeR \cite{DBLP:conf/eccv/RohrbachRHDS16}, MATN  \cite{DBLP:conf/cvpr/0006LZF18}, VC \cite{DBLP:conf/cvpr/ZhangNC18}, KAC Net \cite{DBLP:conf/cvpr/ChenGN18} and IGN \cite{DBLP:conf/nips/ZhangZLZH20}. Then we compare our method with the variant of MAttNet, which combines MattNet and GroundeR to solve the weakly supervised REG problem.

\begin{table*}[]
	\centering
	\centering
	\caption{Accuracy (IoU $>$ 0.5) of EARN with different network modules on RefCOCO, RefCOCO+ and RefCOCOg dataset. `adp', `lan' and `att' denote adaptive reconstruction, language reconstruction and attribute classification, respectively. `ent' means entity enhancement and `scxtp' represents soft context pooling.}
	\scalebox{1.25}{
		\begin{tabular}{c|ccccc|ccc|ccc|c}
			\hline
			\multirow{2}{*}{}&\multicolumn{5}{c|}{Settings}&\multicolumn{3}{c|}{RefCOCO}&\multicolumn{3}{c|}{RefCOCO+}& RefCOCOg\\ 
			\cline{2-13}&adp&lan&att&ent&scxtp&val&testA&testB&val&testA&testB& val\\ \hline 		
			case 1a & $\checkmark$ &&$\checkmark$&& &33.07 & {36.43} & 29.09 & 33.53 & {36.40} & 29.23 & 33.19 \\
			
			
			case 2a & $\checkmark$&$\checkmark$&&&& 33.60 & 35.65& 31.48 & 34.40 & 35.54 & 32.60 & 34.50\\
			
			case 3a &$\checkmark$&$\checkmark$&$\checkmark$&&&{34.26}&{36.01}&{33.07}&{34.53}&{36.01}&{33.75}&{34.66}\\ 
			
			case 4a &&$\checkmark$&$\checkmark$&$\checkmark$&&34.93&33.76&36.98&35.31&33.46&{37.27}&{38.37}\\ 
			
			case 5a &$\checkmark$&$\checkmark$&$\checkmark$&$\checkmark$&&38.45&38.48&38.00&37.60&37.34&36.82		&42.81 \\
			case 6a &$\checkmark$&$\checkmark$&$\checkmark$&$\checkmark$&$\checkmark$& 38.08 & 38.25 & 38.59 & 37.54 & 37.58 & 37.92 & 45.33 \\ 		
			
			\hline
	\end{tabular}}	
	\label{4}
\end{table*}
\begin{table}[!htbp]
	\centering
	\caption{Accuracy (IoU $>$ 0.5) comparison between EARN and the variant of supervised method (MattNet~\cite{DBLP:conf/cvpr/Yu0SYLBB18}+GroundeR)~\cite{DBLP:conf/eccv/RohrbachRHDS16}.  }
	\scalebox{0.98}{
		\begin{tabular}{c|c|cl}
			\hline    
			\small    
			&&MattNet+GroundeR&EARN ($L_{lan}+L_{adp}$)\\ \hline
			\multirow{3}{*}{RefCOCO}&val&21.75&38.08    \space {(+16.33)} \\
			&testA&18.37&38.25  \space {(+19.88)} \\
			&testB&26.36&38.59  \space {(+12.23)} \\  \hline
			\multirow{3}{*}{RefCOCO+}&val&22.06&37.54  \space {(+15.48)} \\
			&testA&18.51&37.58  \space {(+19.07)} \\
			&testB&27.31 &37.92  \space {(+10.61)} \\  \hline
			RefCOCOg&val&28.40 & 45.33 \space {(+16.93)} \\ \hline
			RefCLEF&val&21.86&  36.68 \space {(+14.82)} \\    
			\hline
	\end{tabular}}
	\label{3}
\end{table}

\begin{table}[]
	\centering
	\caption{Accuracy (IoU $>$ 0.5) of EARN with different network modules on RefCLEF dataset. 'adp', 'lan', 'att', 'ent' and 'scxtp' in settings denote adaptive reconstruction, language reconstruction, attribute classification, entity enhancement and soft context pooling respectively.}
	\scalebox{1.1}{	\begin{tabular}{c|ccccc|c}
			\hline
			\multirow{2}{*}{}&\multicolumn{5}{c|}{Settings}& RefCLEF\\ 
			\cline{2-7}&lan&adp&att&ent&scxtp&val\\ \hline 		
			case 1b & $\checkmark$ &&&&&  21.86 \\
			
			case 2b &$\checkmark$&$\checkmark$&&&& 25.35\\
			
			case 3b &$\checkmark$&$\checkmark$&$\checkmark$&&& 26.19\\			
			
			case 4b &$\checkmark$&&$\checkmark$&$\checkmark$&& 33.87\\
			
			
			case 5b &$\checkmark$&$\checkmark$&$\checkmark$&$\checkmark$&$\checkmark$&36.68\\ 	
			\hline
		\end{tabular}
	}
	\label{5}
\end{table}
\subsubsection{Comparison with Weakly Supervised Methods}
We compare our EARN with recent state-of-the-art weakly supervised REG methods on five datasets. 
The LRCN uses image captioning to score how likely the query phrase is to be generated for the proposal box. 
CAFFE-7K predicts a class for each proposal box, and then compares the class with the query phrase after they are projected into a joint vector space. 
GroundeR learns to ground by reconstructing a given phrase using an attention mechanism. 
MATN takes region proposals as anchors and learns a multi-scale correspondence network to search for the referent.
VC uses a variational Bayesian framework to model the combinatorial context configurations.
KAC Net leverages complementary external knowledge to ground the referent.
IGN generates counterfactual positive and negative results to train weakly supervised vision-language grounding model with contrastive learning.
These methods achieve remarkable results on weakly supervised REG.

TABLE \ref{table:coco_acc} shows the results on RefCOCO, RefCOCO+ and RefCOCOg datasets. 
On the RefCOCO dataset, EARN outperforms the state-of-the-art method (CCL) by 0.61\% on testA and 6\% on testB. On the RefCOCO+ dataset, the accuracy of EARN is 0.67\% and 4.36\% higher than the previous method. 
These improvements come from the adaptive grounding and collaborative reconstruction, with which the model can better distinguish the same-category objects.
Besides, we can find that the accuracy of our method on the RefCOCOg dataset is 10.41\% higher than IGN. The performance gain is much higher than that on RefCOCO and RefCOCO+ datasets. This is because the average length of referring expressions in the RefCOCOg dataset is around 8.4, while the RefCOCO and RefCOCO+ datasets are only about 3.6. Longer queries tend to contain more abundant context information. 
EARN can comprehensively utilize the context information with our adaptive grounding to achieve better performance.
Benefiting from the entity enhancement and soft context pooling, the performance of EARN is much better (+ 10.67\% on RefCOCOg) than the previous version ARN. 

We also conduct experiments where only the adaptive reconstruction is reserved and the results are shown in Table  \ref{table:coco_acc}.
Compared with the original network, we can see that on testB of RefCOCO, testB of RefCOCO+, and val of RefCOCOg, the accuracy drops a lot compared with the final results. This is probably because we minimize the distance between the learned visual and language features in the adaptive visual reconstruction. Without directly reconstructing the original, the network may lose some information.
In contrast, we conduct experiments without adaptive reconstruction, and the accuracy drops compared with the final results, especially on testA of RefCOCO, testA of RefCOCO+, and val of RefCOCOg dataset. This may come from that testA contains multiple people and testB contains other objects. Adaptive reconstruction can better handle the situation where same-category objects are situated together.



Table \ref{table:clef_acc} reports the results on RefCLEF, Flickr30k datasets. 
We can observe that the performance of previous method is far less than the performance on the RefCOCO series datasets. This is because the initial number of candidate proposals is much higher than other datasets. With the adaptive grounding and collaborative reconstruction, the previous version of EARN improves its performance by 10\% compared with the KAC net. Further, benefitting from the candidate selection based on entity enhancement, the accuracy is 21\% higher than that of the previous state-of-the-art method. 
The accuracy increment is little on Flickr30k dataset compared with KAC Net. This may come from that we omit the context information on this dataset. {Flickr30k dataset focuses more on phrase grounding, which does not require that each image must contain objects of the same category.} The advantage of EARN to distinguish the target object from other objects of the same category is unnecessary on this dataset. Thus the performance gain is less compared with other datasets.


\begin{table*}[]
	\centering
	\caption{The ablation study of features and entity enhancement on RefCOCO, RefCOCO+ and RefCOCOg datasets. 'attr', 'loc' and 'cxt' denote attribute features, location features and context features, respectively.  'hsf', 'ssf' and 'distp' are short for hard subject filter, soft subject filter and distance penalty. 'adp' and 'scxtp' represent adaptive grounding and soft context pooling.}
	\begin{tabular}{c|ccc|ccc|cc|ccc|ccc|c}
		\hline
		\multirow{2}{*}{}&\multicolumn{8}{c|}{Settings}&\multicolumn{3}{c|}{RefCOCO}&\multicolumn{3}{c|}{RefCOCO+}& RefCOCOg\\ 
		\cline{2-16}&attr&loc&cxt&hsf&ssf&distp&adp&scxtp&val&testA&testB&val&testA&testB& val\\ \hline 		
		case 1c& $\checkmark$ &&&&&&& &15.58&8.34&25.57&16.25&14.15&18.41&33.64 \\
		
		case 2c&$\checkmark$&$\checkmark$&&&&&&& 36.47&35.11&37.74&20.33&17.34&24.38&36.13\\
		
		case 3c& $\checkmark$&$\checkmark$&$\checkmark$&&&&&& 36.73&35.60&37.11&36.49&35.92&38.41&37.17\\
		
		case 4c &$\checkmark$&$\checkmark$&$\checkmark$&$\checkmark$&&&&&35.31&33.87&36.17&34.78&33.34&37.43&40.24\\ 
		
		case 5c &$\checkmark$&$\checkmark$&$\checkmark$&&$\checkmark$&&&&35.28&35.41&36.60&35.36&35.16&36.29&38.45\\ 
		
		case 6c &$\checkmark$&$\checkmark$&$\checkmark$&$\checkmark$&&$\checkmark$&&&35.98&35.28&37.39&35.66&33.67&38.52&43.16\\
		
		case 7c &$\checkmark$&$\checkmark$&$\checkmark$&&$\checkmark$&$\checkmark$&&&{36.34}&{35.28}&{37.72}&{37.16}&{36.06}&{39.29}&36.65\\
		
		case 8c &$\checkmark$&$\checkmark$&$\checkmark$&$\checkmark$&&$\checkmark$&$\checkmark$&& 38.45&38.48&38.00&37.60&37.34&36.82		&42.81 \\ 	
		case 9c &$\checkmark$&$\checkmark$&$\checkmark$&$\checkmark$&&&$\checkmark$&$\checkmark$& 38.08 & 38.25 & 38.59 & 37.54 & 37.58 & 37.92 & 45.33  \\ 	
		
		\hline
		
	\end{tabular}
	\label{6}
\end{table*}

\begin{table}[]
	\centering
	\caption{Ablation study of different loss proportion on RefCLEF dataset. $\alpha$, $\beta$, $\gamma$, $\lambda$ denote the weights on $Loss_{avis}$, $Loss_{alan}$, $Loss_{lan}$, $Loss_{att}$, respectively. }
	\scalebox{1}{\begin{tabular}{c|cccc|c}
			\hline
			& $\alpha$&$\beta$&$\gamma$&$\lambda$&val \\ \hline
			case 1d & 0.001&1&10&0&24.14\\
			case 2d & 0.01 &1 &10&0&21.83\\ 
			case 3d & 0.001 &10 &10&0&22.55\\
			\hline
			case 4d & 0.001&1&1&0&22.34\\ 
			case 5d& 0.001&1&1&1&25.35\\ \hline
			
			case 6d & 0.001&1&10&1&24.34\\ 
			case 7d& 0.001&1&20&1&24.76\\ 
			case 8d& 0.001&1&30&1&26.19\\
			case 9d&0.001&1&40&1&25.53\\
			\hline
			case 10d&0&0&1&0&21.86\\

			\hline
	\end{tabular}}
	
	\label{table:clef_loss}
\end{table}

\subsubsection{Comparison with Variants of Supervised Method}
We compare our method with the variant of MAttNet~\cite{DBLP:conf/cvpr/Yu0SYLBB18} in this section. MAttNet is one of the most popular networks in supervised REG and gets promising grounding results.
With the language reconstruction loss~\cite{DBLP:conf/eccv/RohrbachRHDS16} instead of the hinge loss, MattNet can deal with the weakly supervised REG problem.
Table \ref{3} shows the comparison results, and we can find our method improves the performance with a large margin on all four datasets. 
The improvement mainly comes from the entity enhancement and adaptive reconstruction. The entity enhancement helps to discard the most unrelated proposals. The adaptive reconstruction is vital to capture the key and discriminative component in the query to distinguish the target object from other objects of the same category. The disentangled mapping also brings stronger interpretation between query and proposal.

\begin{table*}[t]
	\centering
	\caption{Comparison between parsing-based and attention-learned language features. }
	\begin{tabular}{c|ccc|ccc|c}
		\hline
		\multirow{2}{*}{Methods}&\multicolumn{3}{c|}{RefCOCO}&\multicolumn{3}{c|}{RefCOCO+}& RefCOCOg\\ 
		\cline{2-8}&val&testA&testB&val&testA&testB& val\\ \hline 
		parsing-based &38.28&37.33&38.27&36.93&37.22&36.65&44.96 \\ 
		attention-learned & 38.08& 38.25 & 38.59 & 37.54 & 37.58& 37.92&45.33  \\ 					
		\hline	
	\end{tabular}
	\label{table:parsed_lang}
\end{table*}
	
	We also conduct the experiment to compare parsing-based and attention-based language features.
	For parsing-based language feature learning, we first use the off-the-shelf language parser to parse the referring expression into subject, location and context. {Then the parsed components are directly fed into the LSTM network to learn the corresponding language features.}  	Table~\ref{table:parsed_lang} shows the comparison results. Overall, the accuracy slightly degrades compared with attention-based method. 
	This phenomenon mainly results from the pre-trained language parser network's complex language syntax and upper bound. 

\subsection{Ablation Study}
We conduct experiments to evaluate the impact of different network structures and parameters. Ablation studies are implemented on entity enhancement, adaptive reconstruction, language reconstruction, attribute classification and context modeling. We also perform experiments of different filter thresholds and loss proportions in this section.

\subsubsection{Network Modules}
Table \ref{4} presents the results of EARN with different network modules on RefCOCO, RefCOCO+ and RefCOCOg datasets and Table \ref{5} shows that on RefCLEF dataset.  
'adp', 'lan', 'att', 'ent' and 'scxtp' in settings denote adaptive reconstruction, language reconstruction, attribute classification, entity enhancement and soft context pooling respectively. We can observe several findings.

{First, the comparison of case 1a, case 2a and case 3a, case 1b, case 2b and case 3b shows that the collaborative reconstruction can perform better than single or the combination of two reconstruction methods.} The adaptive reconstruction helps to capture the discriminative component to distinguish the target from other same-category objects. The language reconstruction ensures the model does not ignore any informative language information. The attribute classification further enhances attribute learning. Each of them plays a vital role in grounding a particular target. 

{Second, the composition of entity enhancement and language reconstruction (case 4a and case 4b) improves the performance by a large margin. }
The entity enhancement introduces semantic similarity between parsed expression and extracted proposal category
as supervision, thus guiding the selection of candidate proposals and excluding the unrelated proposals. 
It leverages meaningful information to alleviate the lack of region-level annotations.

{Third, compared with case 3a and case 4a, the results in case 5a show that the collaborative reconstruction can further promote the performance of the grounding.} The collaborative reconstruction adaptively learns the correspondences between visual and language features upon subject, location and context. With the discriminative location and context information, EARN can distinguish a particular object from other same-category objects, thus handling the situation with multiple objects of the same category.

	\begin{table*}[t]
		\centering
		\caption{Comparison of different context pooling. `5cxtp' represents the max-context pooling of 5 surrounding proposals, `mcxtp' is short for max-context pooling, and `scxtp' denotes soft-context pooling. }
		\begin{tabular}{c|ccc|ccc|c}
			\hline
			\multirow{2}{*}{Methods}&\multicolumn{3}{c|}{RefCOCO}&\multicolumn{3}{c|}{RefCOCO+}& RefCOCOg\\ 
			\cline{2-8}&val&testA&testB&val&testA&testB& val\\ \hline 
			 5cxtp &34.26&36.01&33.07&34.53&36.01&33.75&34.66 \\ 
			mcxtp& 38.45& 38.48 & 38.00 & 37.60 & 37.34& 36.82&42.81  \\ 			
			scxtp& 38.08& 38.25 & 38.59 & 37.54 & 37.58& 37.92&45.33  \\ 			
			\hline	
		\end{tabular}
		\label{table:comparison_on_cxtp}
\end{table*}

\begin{table}[t]
		\centering
		\caption{Accuracy of soft context pooling on complex relationship. `mcxtp' is short for max-context pooling, and `scxtp' denotes soft-context pooling. }
		\begin{tabular}{c|c|c|c}
			\hline
			Methods&RefCOCO&RefCOCO+& RefCOCOg\\ \hline 
			mcxtp&17.46&20.88&43.11 \\ 
			scxtp & 21.75& 21.66 & 46.47 \\ \hline
			num	&	653		&	637		& 4233
			\\ 					
			\hline	
		\end{tabular}
		\label{table:scxtp_on_complex}
\end{table}

	We also compare different context pooling methods in Table~\ref{table:comparison_on_cxtp}. `5cxtp' shows the version without entity enhancement, where we select the proposal with the maximum response to the query from 5 surrounding candidate proposals as object.
	In `mcxtp', we select the proposal with the maximum score from all the candidate proposals as object based on entity enhancement. 
	'scxtp' shows the results with soft context pooling, where the context is modeled based on the features of all candidate proposals. The weights are the learned object matching scores. 
	We can observe that 'scxtp' achieves better results on most datasets.
	Compared with the max context pooling, soft context pooling encodes context features by learning all the candidate proposals in the image. It can better model the multi-entity relationship. 

\begin{figure}[tp]
	\centering
	\includegraphics[width=0.5\textwidth]{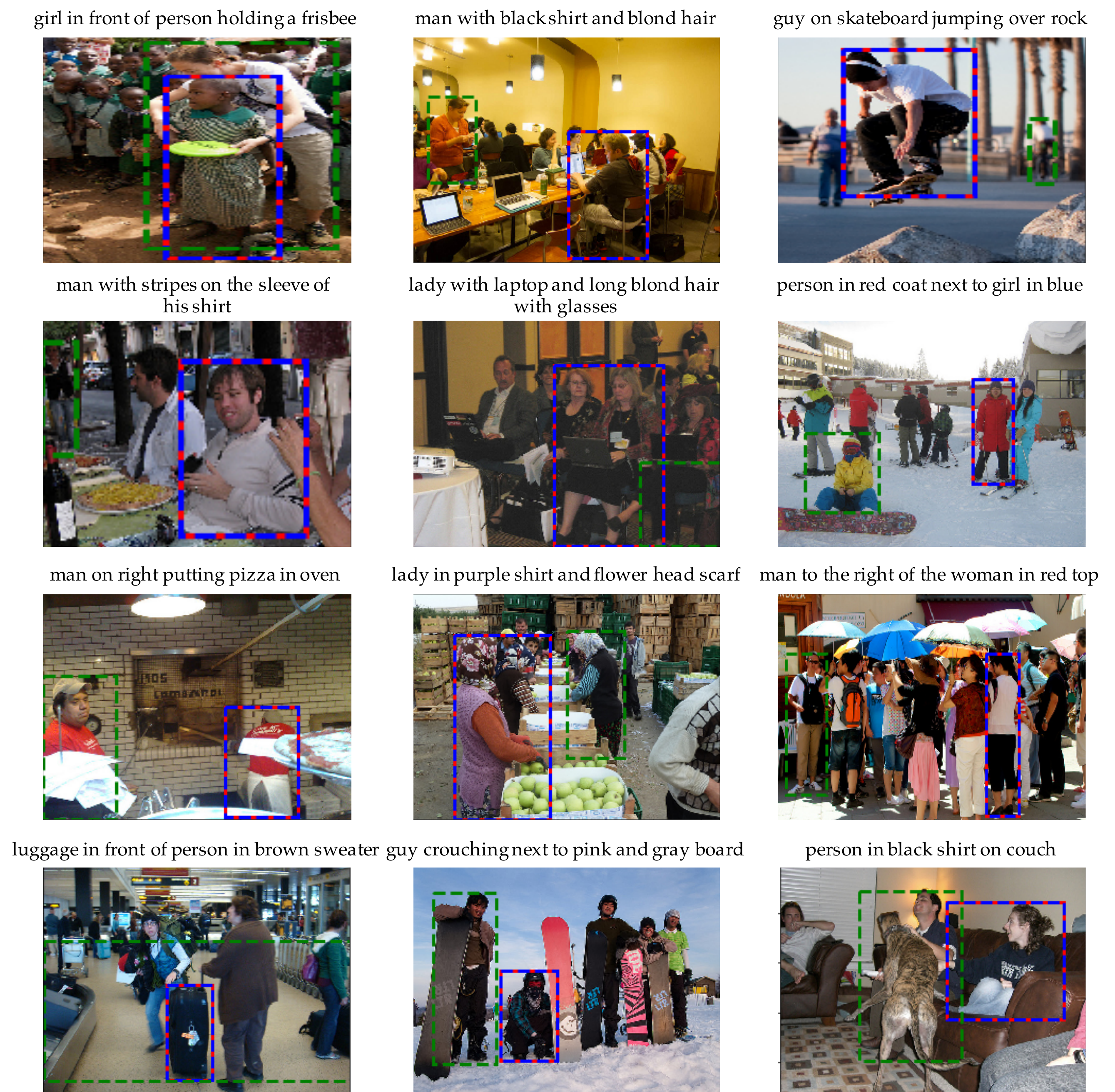}
	\caption{Qualitative results on referring expression with complex relationship. The denotations of the bounding box colors are as follows. Solid red: ground truth; dashed blue: predicted proposal with soft context pooling; dashed green: predicted proposal with max context pooling. }
	\label{fig:scxtp_on_complex}
\end{figure}

	To verify our method on dealing with more complex relationships, we collect the referring expressions with complex relationships from the validation and test set of RefCOCO, RefCOCO+ and RefCOCOg (mainly based on query length). 
	The comparison results are shown in Table~\ref{table:scxtp_on_complex}. Compared with the max context pooling, soft context pooling can achieve better results on referring expression grounding with more complex relationships, which increases by 4.29\%, 0.78\%, and 3.36\% on RefCOCO, RefCOCO+, and RefCOCOg, respectively. 
	As shown in Fig.~\ref{fig:scxtp_on_complex}, we also present some qualitative results of soft context pooling and max context pooling on referring expression with complex relationships. We can see that EARN with soft-context pooling can better distinguish the target when a complex relationship exists. 
	{We provide our evaluation set for complex relationship cases online~\cite{github_earn}.}

	\begin{table*}[]
		\centering
		\setlength{\tabcolsep}{3.5mm}
		\caption{Albation study of different loss propotion on RefCOCO, RefCOCO+ and RefCOCOg datasets. $\alpha$, $\beta$, $\gamma$, $\lambda$ denote the weights on $Loss_{avis}$, $Loss_{alan}$, $Loss_{lan}$, $Loss_{att}$, respectively. }
		\scalebox{1}{
			\begin{tabular}{c|cccc|ccc|ccc|c|c}
				\hline
				\multirow{2}{*}{}&\multicolumn{4}{c|}{Settings}&\multicolumn{3}{c|}{RefCOCO}&\multicolumn{3}{c|}{RefCOCO+}& RefCOCOg&\\ 
				\cline{2-13}&$\alpha$&$\beta$&$\gamma$&$\lambda$&val&testA&testB&val&testA&testB&val&sum\\ \hline 
				
				case 1e  &  1&1  &1  &0   &32.92  & \textbf{36.40} &29.26&33.06&36.34&29.60&33.08 & 230.66\\
				
				case 2e &   0.01&1  &1  &1   &\textbf{34.32} &36.24&33.05&\textbf{35.60}&\textbf{36.92}&33.09&34.44&\textbf{243.66} \\
				
				case 3e & 0.01&1&5&1&34.26&36.01&33.07&34.53&36.01&33.75&34.66&242.29\\
				case 4e& 0.01&1&10&1&34.18&35.83&32.29&32.39&33.39&32.89&34.24&235.21\\
				case 5e& 0.01&1&15&1&29.09&27.13&\textbf{33.09}&29.97&27.98&33.99&\textbf{34.94}&216.19\\
				case 6e&0.01&1&20&1&29.87&27.86&33.05&29.20&25.57&\textbf{35.28}&34.60&215.43\\
				\hline		
		\end{tabular}}
		\label{table:coco_loss}
\end{table*}
\subsubsection{Features and Entity Enhancement}
We conduct experiments in Table~\ref{6} to evaluate the impact of attribute ('attr'), location ('loc') and context ('cxt') features. 
{From the comparison between case 1c and case 2c, we find that the accuracy improves by a large margin on RefCOCO and RefCOCOg datasets after aggregating the location features.} However, the improvement on RefCOCO+ is less. This may come from that this dataset disallows to use locations.
{Then, we compare case 2c with case 3c. The results show that context features can further improve the performance, especially on the RefCOCO+ dataset. }
These findings indicate that the features have different impacts on the datasets.
Adaptively using the features is beneficial to grounding different queries due to their unique distribution.

	Besides, we also try to find a better way to use the calculated supervision. The supervision in the form of semantic similarity to learn the entity attention. After we learn the subject attention score, we design two methods to obtain candidate proposals. 
	Soft filter assigns different weights to the candidate proposals according to the scores. Specifically,  soft filter multiples the final ranking score $S_t^i $ with the subject attention score $Score_s^i$. 
	Different from soft filter, hard filter just discards the proposals under the selected threshold. 
	We put forward a distance penalty to penalize the object attention score if the distance between the target and the related object is large. The distance penalty only exists in max context pooling.
	Table~\ref{6} shows the ablation study on hard subject filter ('hsf') and soft subject filter ('ssf' ) and distance penalty ('distp'). 
	{The comparisons between case 4c and 6c, case 5c and 7c demonstrate that hard filter can better select candidate proposals, and distance penalty can improve the performance to a certain degree. }
	This is because the hard filter can eliminate the noisy information of all the unrelated proposals and the distance penalty conforms to the habit of humans to describe a target with its surrounding object.

We add adaptive grounding ('adp') and soft context pooling ('scxtp') to our best setting of features and enhancement method in Table~\ref{6} and get better results.
Adaptive reconstruction can further improve the accuracy with its disentangled mapping between subject, location and context. 
Compared with max context pooling, soft pooling can handle more complicated context as it considers all the candidate proposals in context modeling.

		\begin{table*}[t]
		\centering
		\caption{Comparison between different context location features. `cxt\_abs' represents the absolute location features, `cxt\_rel' is short for relative location features, and `cat\_all' denotes the combination of absolute and relative location features.}
		\begin{tabular}{c|ccc|ccc|c}
			\hline
			\multirow{2}{*}{Methods}&\multicolumn{3}{c|}{RefCOCO}&\multicolumn{3}{c|}{RefCOCO+}& RefCOCOg\\ 
			\cline{2-8}&val&testA&testB&val&testA&testB& val\\ \hline 
			cxt\_abs &37.98&38.22&37.29&\textbf{37.68}&36.83&37.25&44.11 \\ 
			cxt\_rel& \textbf{38.08}& \textbf{38.25} & \textbf{38.59} & 37.54 & 37.58& \textbf{37.92}&\textbf{45.33 } \\ 		
			cat\_all&37.76&38.02&37.29&37.48&\textbf{38.00}&36.43&44.18  \\ 			
			\hline	
		\end{tabular}
		\label{table:comparison_on_cxtloc}
\end{table*}
	We conduct experiments with different context location features encoding methods. The results are shown in Table~\ref{table:comparison_on_cxtloc}. `cxt\_abs' shows the result the network with absolute context location features, encoded as  $r_l^j = \left[ \frac { x _ { t l } } { W } , \frac { y _ { t l } } { H } , \frac { x _ { b r } } { W } , \frac { y _ { b r } } { H } , \frac { w \cdot h } { W \cdot H } \right]$. `cxt\_rel' demonstrates the performance of the original relative location features. `cxt\_all' presents the accuracy of the combination of absolute and relative location features. From Table~\ref{table:comparison_on_cxtloc}, we can see that relative location features achieves the best performance on most datasets. We suppose that the relative location features can better express the distribution like `A on the left of B'.
	The absolute location features may misguide the grounding of the context proposal. For example, the context proposal may be on the right, but it is on the left of the target. In that circumstance, relative location features can better describe the distribution of the proposals.

\begin{table*}[t]
		\centering
		\caption{Comparison between end-to-end and two-stage training strategy.}
		\begin{tabular}{c|ccc|ccc|c}
			\hline
			\multirow{2}{*}{Methods}&\multicolumn{3}{c|}{RefCOCO}&\multicolumn{3}{c|}{RefCOCO+}& RefCOCOg\\ 
			\cline{2-8}&val&testA&testB&val&testA&testB& val\\ \hline 
			end-to-end training &37.41&38.20&35.98&37.82&36.73&38.52&44.77 \\ 
			two-stage training & 38.08& 38.25 & 38.59 & 37.54 & 37.58& 37.92&45.33\\ 					
			\hline	
		\end{tabular}
		\label{table:end2end}
\end{table*}
	We also conduct experiments including visual encoder fine-tuning, which is in an end-to-end manner. The visual encoder is fine-tuned with the SGD algorithm with a learning rate of 1e-5. The other parts of EARN are trained through Adam with an initial learning rate of 4e-4. The experimental results of end-to-end training are shown in Table~\ref{table:end2end}. On the val and testB of the RefCOCO+ dataset, the accuracy of the end-to-end training strategy is slightly better than the two-stage strategy. 
	The two-stage training achieves a higher performance on the other datasets, especially the testB of RefCOCO.
	This is perhaps because the fixed visual features can better learn the correspondence between the two modalities upon the subject, location and context in EARN. 
	We find it necessary to keep either language or visual features fixed to learn their correspondence because there is no ground-truth.
	Sometimes a simple linear layer may destroy the training since the network parameters might be set to zero roughly to reach convergence as soon as possible. 
	If the visual and language features update simultaneously, it becomes difficult for the network to learn the disentangled mapping. 
	It takes 8.3 hours to train the network end-to-end with a GTX 1080Ti, while it takes 6.3 hours to train the two-stage network.

\subsubsection{Network Hyperparameters}
\begin{figure}[t]  
	
		
	\subfigure{
		\includegraphics[width=0.306\columnwidth]{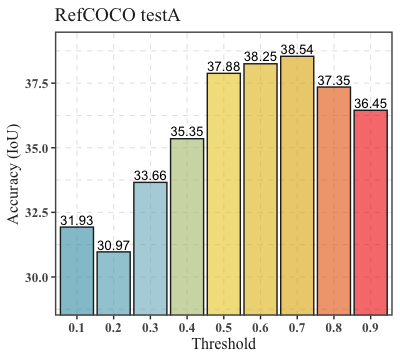} 
	}
	\subfigure{
		\includegraphics[width=0.306\columnwidth]{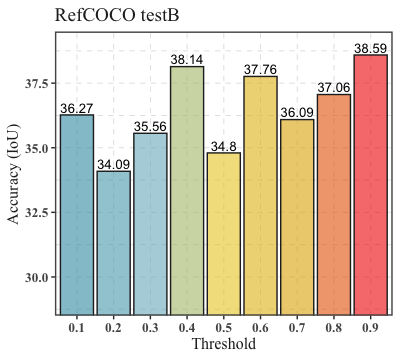} 
		
	}
	\subfigure{
		\includegraphics[width=0.306\columnwidth]{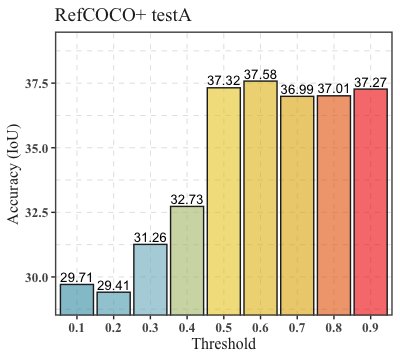} 
	}

	\subfigure{
		\includegraphics[width=0.306\columnwidth]{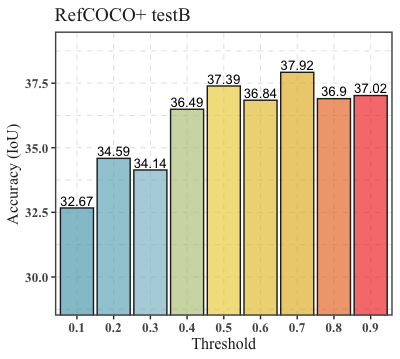} 
	}
	\subfigure{
		\includegraphics[width=0.306\columnwidth]{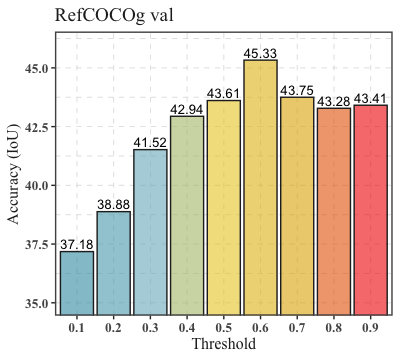} 
	}
	\subfigure{
		\includegraphics[width=0.306\columnwidth]{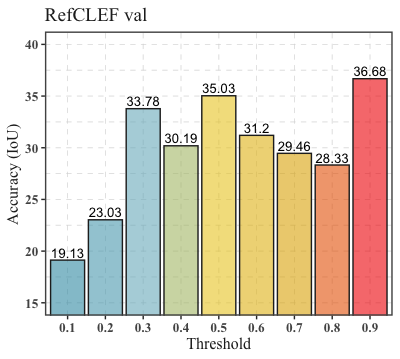} 
	}
	\caption{\label{fig:sparsity} Accuracy (IoU) of different threshold for hard subject filter on RefCOCO, RefCOCO+, RefCOCOg and RefCLEF.}
\end{figure}

\begin{figure*}[tp]
	\centering
	\includegraphics[width=0.8\textwidth]{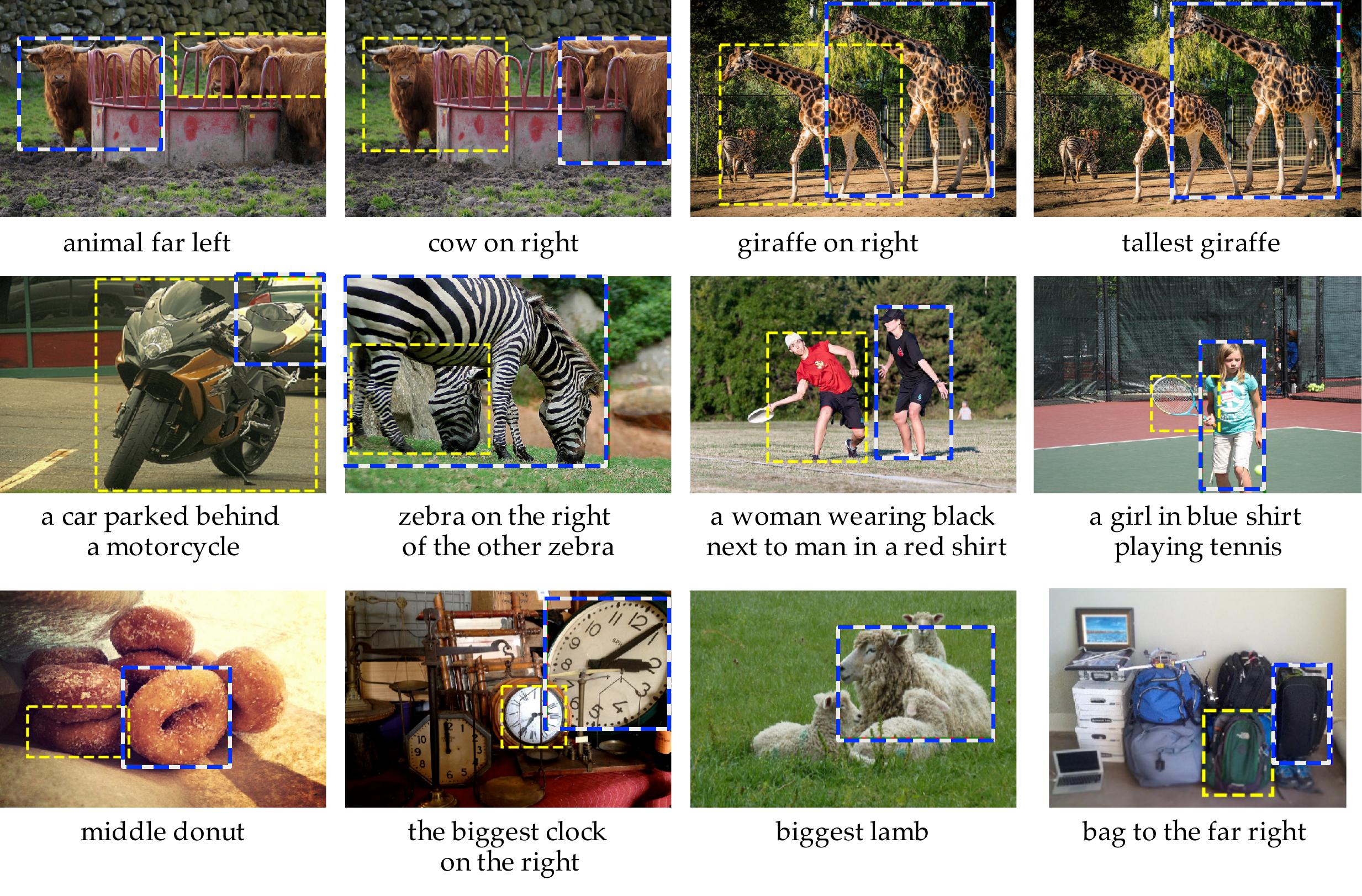}
	\caption{Qualitative results on RefCOCO, RefCOCO+ and RefCOCOg datasets. The denotations of the bounding box colors are as follows. Solid white: ground truth; dashed blue: predicted proposal; dashed yellow: context ground. }
	\label{visualization}
\end{figure*}

In this section, we first perform experiments to find the best threshold for hard subject filter. Fig. \ref{fig:sparsity} shows the results of different threshold on RefCOCO, RefCOCO+, RefCOCOg and RefCLEF. We can find that if we discard candidate proposals whose attention scores are under 0.6, we can achieve the best result in most cases. A threshold of 0.6 enables the model to remain the target proposal with bigger probability as well as discard the noisy information. 

Second, we study the proportion of each loss in the collaborative reconstruction by running ablation experiments. 
Table \ref{table:clef_loss} shows the results of different loss proportion on RefCLEF.
$\alpha$, $\beta$, $\gamma$, $\lambda$ denote the weights on $Loss_{avis}$, $Loss_{alan}$, $Loss_{lan}$, $Loss_{att}$, respectively. $Loss_{adp}$ is composed of $Loss_{avis}$ and $Loss_{alan}$.
The adaptive visual reconstruction loss is first set as 0.001 based on the order of magnitude. {We try different proportions of $Loss_{avis}$ and $Loss_{alan}$ in case 2d and case 3d compared to case 1d, respectively.} We find the result is better when $\alpha$ is 0.001, due to the order of magnitude in $Loss_{avis}$. {The comparison of case 4d and case 5d shows that attribute classification loss is beneficial to the grounding results. Case 6d, case 7d, case 8d and case 9d show that the performance of the network is higher when the proportion $\lambda$ of $Loss_{lan}$ is 30. However, when we only use the $Loss_{lan}$ in case 10d, the results are not as good as the combination of $Loss_{adp}$ and $Loss_{lan}$.} We can conclude that adaptive reconstruction is necessary to discriminate the target with the same-category objects. 

\begin{figure*}[tp]
	\centering
	\includegraphics[width=0.85\textwidth]{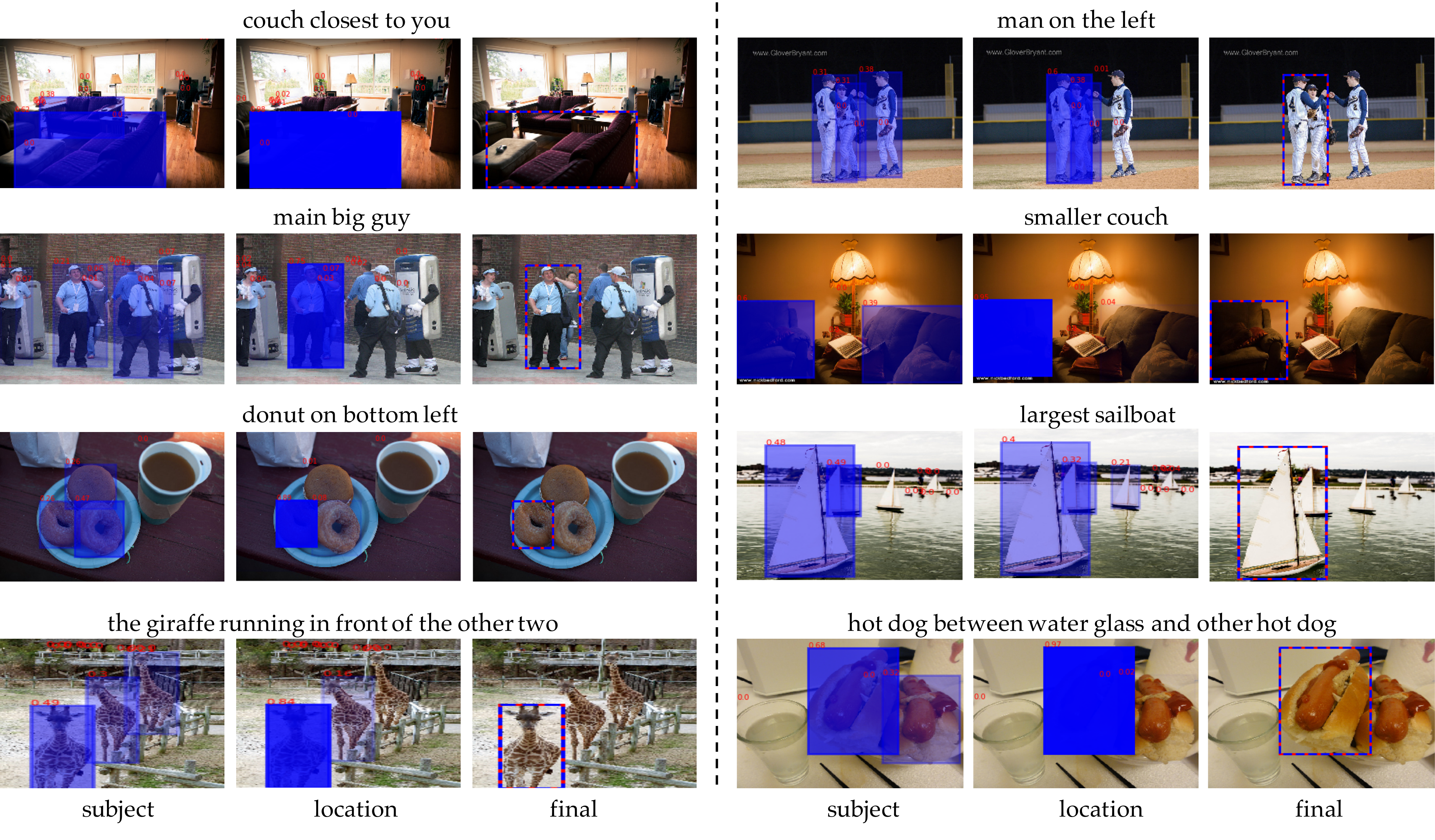}
	\caption{More detailed qualitative results on RefCOCO, RefCOCO+ and RefCOCOg datasets. The denotations of the bounding box colors are as follows. Solid red: ground truth; dashed blue: predicted proposal. The saturation of the blue boxes denotes the attention score of each proposals.}
	\label{vis_detail}
\end{figure*}

\begin{figure}[tp]
	\centering
	\includegraphics[width=0.44\textwidth]{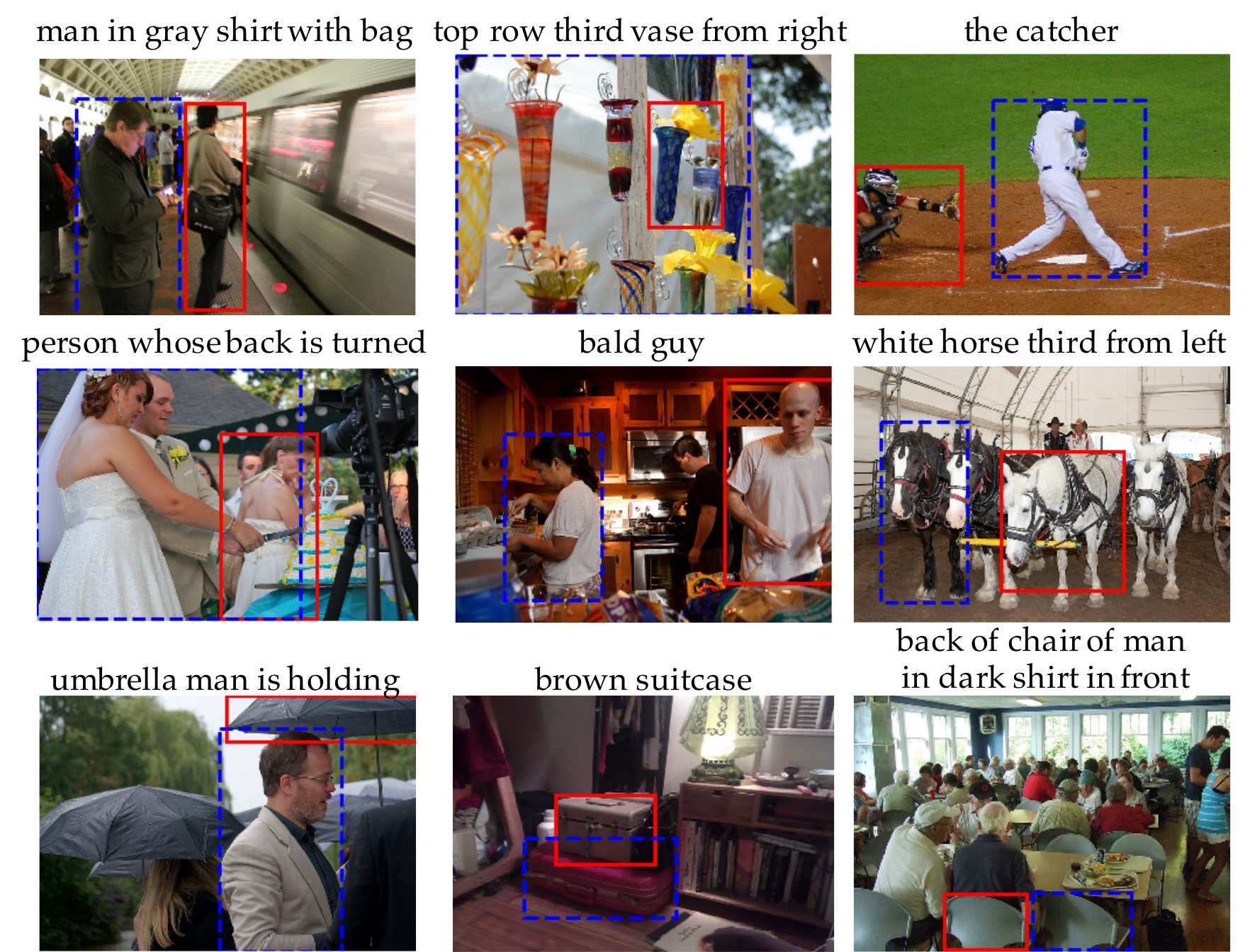}
	\caption{Some failure cases on RefCOCO, RefCOCO+ and RefCOCOg datasets. The denotations of the bounding box colors are as follows. Solid red: ground truth; dashed blue: predicted proposal. }
	\label{failure}
\end{figure}

Table~\ref{table:coco_loss} reports the results of EARN with different proportions of each loss in the collaborative reconstruction on RefCOCO, RefCOCO+ and RefCOCOg.  
We find that when $Loss_{lan}$ ($\gamma$) accounts for a more significant part in the collaborative loss, the performance on testA will drop greatly. When the proportion of $Loss_{adp}$ (case 1e) is bigger, the results in testB are worse. This may result from that testA only contains person while testB consists of multiple categories. In testA, how to distinguish different persons is vital so adaptive reconstruction is necessary. In testB, the category is informative to ground the particular target.
After the parameter search, based on the summation of all results on each dataset in Table~\ref{table:coco_loss}, we find that the settings in case 2e achieve more balanced results on all the datasets. 

\subsection{Qualitative Results}

Fig. \ref{visualization} presents some qualitative examples on RefCOCO, RefCOCO+ and RefCOCOg datasets. The query is shown below the corresponding images. 
The first row shows the result based on different queries in the same image. The proposed EARN can handle the location information correctly. The second row shows some examples with context information. EARN correctly grounds both the referential object and context object. The third row shows some difficult examples where multiple objects of the same category exist. It shows that the proposed EARN can help ground the hard cases that contain multiple objects of the same category. This may benefit from the adaptive grounding and reconstruction, which focus on adaptively learning the subject, location and context information.

Fig. \ref{vis_detail} shows more detailed qualitative results. We use saturation to represent the attention score. The columns of 'subject' and 'location' show the final ranking score for the candidate proposals upon subject and location. 'final' denotes the final grounding result. We can see that EARN grounds the final result based on the different attention modules. Taking ``donut on bottom left'' as an example, in the subject ranking score, we can hardly find the left donut. However, with location attention, the model can easily ground the bottom left donut. 

Fig. \ref{failure} shows some failure cases on RefCOCO, RefCOCO+ and RefCOCOg datasets. The queries like ``top row third vase from right'' and ``white horse third from left'' are hard to understand. Although the model can learn 'right' and 'left',  it cannot count 'third'. 
Besides, as we do not have the region-level ground-truth, our method learns the mapping between language and region proposal based on the whole dataset. This results in that it is difficult to recognize the rare words in the datasets, such as 'catcher', 'bold'. and the uncommon colors like 'brown'.
EARN can be confused when coming across to the complex expressions, such as ``umbrella man is holding'', ``person whose back is turned'', ``back of chair of man in dark shirt front''. These failure cases can potentially be fixed by applying a complex language extractor.

\section{Conclusion}
We propose an entity-enhanced adaptive reconstruction network to handle the weakly supervised referring expression grounding problem. 
EARN learns to map the correspondence between the image proposals and the language query adaptively with entity enhancement. 
Firstly, we calculate the semantic similarity as supervision to discard the unrelated candidate proposals in entity enhancement module. 
Secondly,  we calculate the ranking score for the candidate proposals upon the subject, location and context in the adaptive grounding module. 
Finally, we design collaborative reconstruction loss, including language reconstruction loss, adaptive reconstruction loss and attribute classification loss to measure the grounding performance.
Experiments conducted on five datasets show the effectiveness of our method. 
We find that our model is not sensitive to counting, less frequent words, and a more complex syntax from the failure cases. 
We will focus more on the above problems to learn more robust weakly supervised REG models in future work.


\ifCLASSOPTIONcompsoc
  \section*{Acknowledgments}
\else
  \section*{Acknowledgment}
\fi

The authors would like to thank the associate editor and the reviewers for their time and effort provided to review the manuscript. This work was supported by the National Key R\&D Program of China under Grant 2018AAA0102003.

\ifCLASSOPTIONcaptionsoff
  \newpage
\fi



\bibliographystyle{IEEEtran}
\bibliography{IEEEabrv,bare_jrnl}
%



%

\newpage
\begin{IEEEbiography}[{\includegraphics[width=1in,height=1.25in,clip,keepaspectratio]{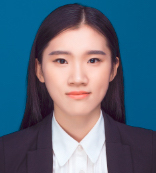}}]{Xuejing Liu}
	received the B.S. degree from Wuhan University in 2016. She is currently pursuing the Ph.D. degree with the Institute of Computing Technology, Chinese Academy of Sciences. She is also with the Key Laboratory of Intelligent Information Processing, Chinese Academy of Sciences. Her current research interests include machine learning, deep learning and computer vision.
\end{IEEEbiography}

\begin{IEEEbiography}[{\includegraphics[width=1in,height=1.25in,clip,keepaspectratio]{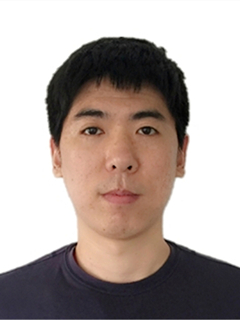}}]{Liang Li}
	received his B.S. degree from Xi’an Jiaotong Univerisity in 2008, and Ph.D degree from Institute of Computing Technology(ICT), Chinese Academy of Sciences(CAS), Beijing, China in 2013. From 2013 to 2015, he held a post-doc position with the Department of Computer and Control Engineering, University of Chinese Academy of Sciences, Beijing, China. Currently he is serving as the associate professor at ICT, CAS. He has also served on a number of committees of international journals and conferences. Dr. Li has published over 80 refereed journal/conference papers, such as IEEE TPAMI, TIP, TNNLS, TMM, CVPR, ICCV, ACM MM, etc. His research interests include image semantic understanding, multimedia content analysis, computer vision.
\end{IEEEbiography}


\begin{IEEEbiography}[{\includegraphics[width=1in,height=1.25in,clip,keepaspectratio]{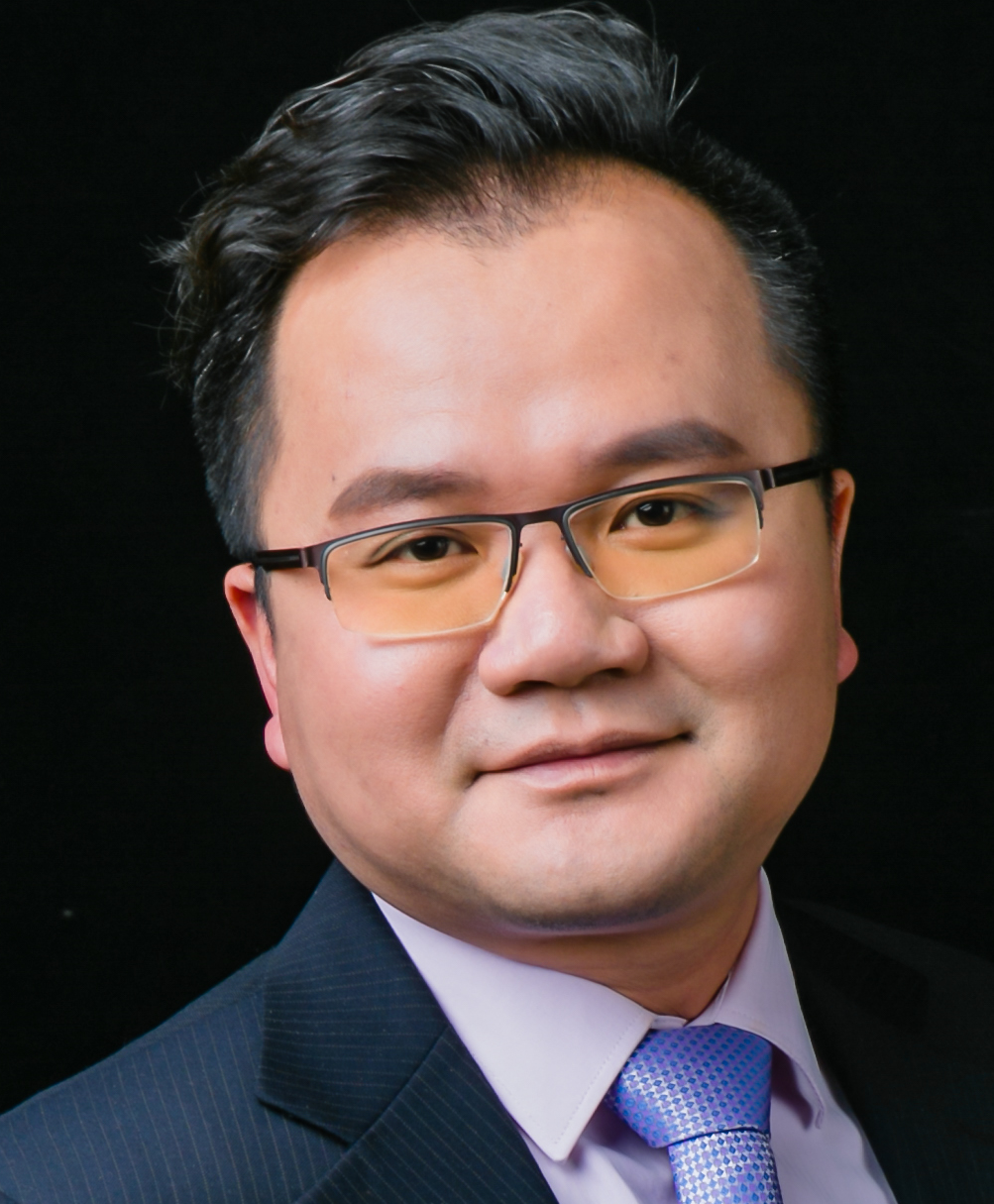}}]{Shuhui Wang} received the B.S. degree in electronics
	engineering from Tsinghua University, Beijing, China, in 2006, and the Ph.D. degree from the Institute of Computing Technology, Chinese Academy
	of Sciences, Beijing, China, in 2012. He is currently a Full Professor with the Institute of Computing Technology, Chinese Academy of Sciences.
	He is also with the Key Laboratory of Intelligent Information Processing, Chinese Academy of Sciences. His research interests include image/video understanding/retrieval, cross-media analysis and visual-textual knowledge extraction.
\end{IEEEbiography}

\begin{IEEEbiography}[{\includegraphics[width=1in,height=1.25in,clip,keepaspectratio]{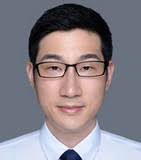}}]{Zheng-Jun Zha}
	received the B.E. and Ph.D. degrees from the University of Science and Technology of China(USTC), Hefei, China, in 2004 and 2009, respectively. He is currently a Full Professor with the School of Information Science and Technology, USTC, the Vice Director of National Engineering Laboratory for Brain-Inspired Intelligence Technology and Application. He was a Researcher with the Hefei Institutes of Physical Science, CAS, from 2013 to 2015, a Senior Research Fellow with the School of Computing, National University of Singapore (NUS), from 2011 to 2013. He has published more than 100 papers in these areas with a series of publications on top journals and conferences. His research interests include multimedia analysis, retrieval and applications, as well as computer vision etc.
\end{IEEEbiography}

\begin{IEEEbiography}[{\includegraphics[width=1in,height=1.25in,clip,keepaspectratio]{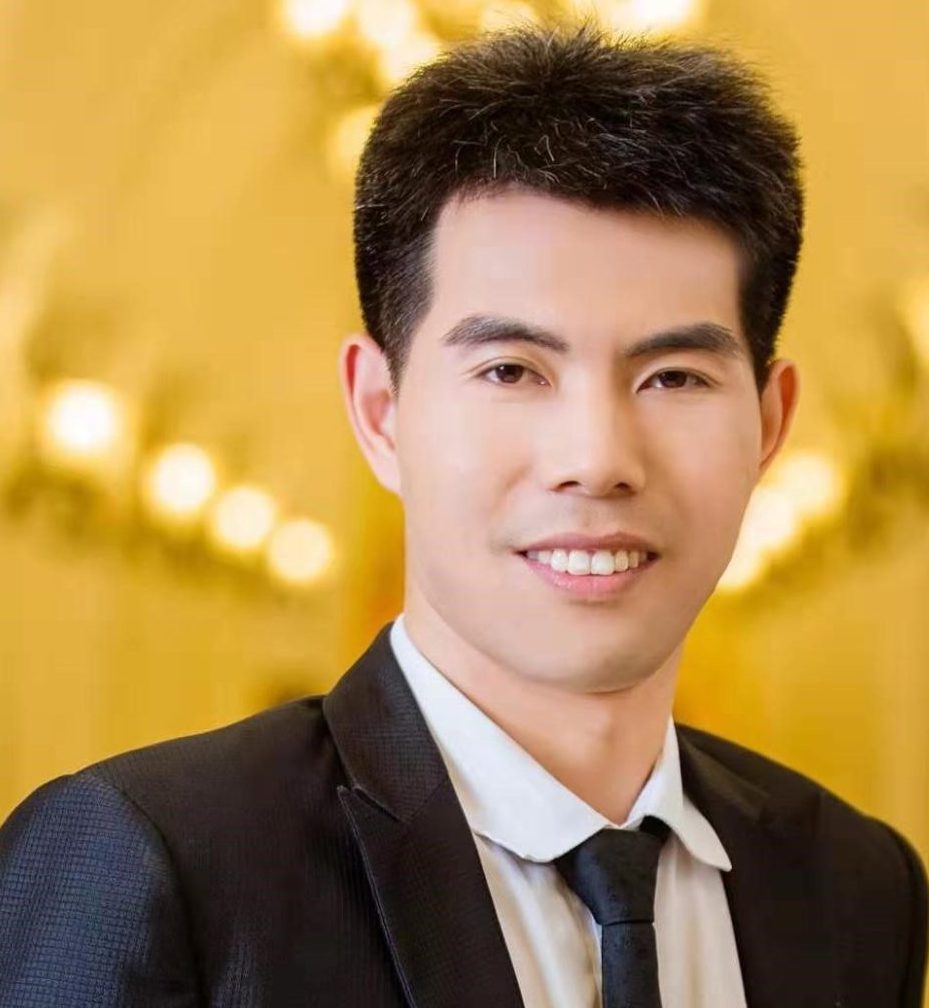}}]{Zechao Li} is currently a Professor at the Nanjing University of Science and Technology. He received his Ph.D degree from National Laboratory of Pattern Recognition, Institute of Automation, Chinese Academy of Sciences in 2013, and his B.E. degree from the University of Science and Technology of China in 2008. His research interests include big media analysis, computer vision, etc. He has authored over 70 articles in top-tier journals and conferences. He serves as an Associate Editor for IEEE TNNLS. He was a recipient of the best paper award in ACM Multimedia Asia 2020, and the best student paper award in ICIMCS 2018.
\end{IEEEbiography}

\begin{IEEEbiography}[{\includegraphics[width=1in,height=1.25in,clip,keepaspectratio]{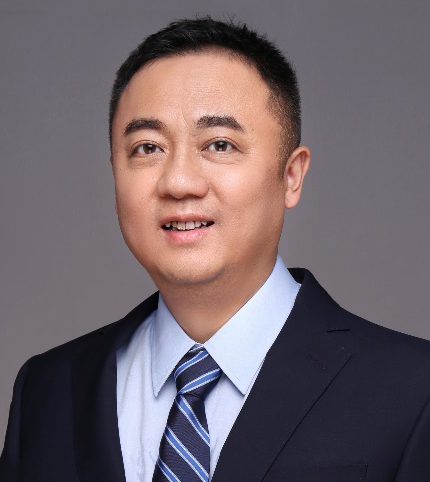}}]{Qi Tian} is currently the Chief Scientist in Artificial Intelligence at Huawei Cloud \& AI. He was the Chief Scientist in Computer Vision at Huawei Noah’s Ark Laboratory from 2018-2020. Before he joined Huawei, Dr. Tian was a Full Professor in the Department of Computer Science, the University of Texas at San Antonio (UTSA). He is an IEEE fellow (2016), IEAS Academician (2021), a Changjiang Chaired Professor of the Ministry of Education, an Oversea Expert by the Chinese Academy of Science. 
Dr. Tian received his Ph.D. in ECE from University of Illinois at Urbana-Champaign (UIUC) in 2002 and received his B.E. in Electronic Engineering from Tsinghua University in 1992 and M.S. in ECE from Drexel University in 1996, respectively. Dr. Tian’s research interests include computer vision, multimedia content analysis, image and video indexing and retrieval, and machine learning, and published over 650 refereed journal and conference papers (including 200 IEEE/ACM Transactions and 210 CCF A Category Conference Papers). His Google citations are over 40000 with h-index of 94. He was the co-author of eight best papers. 
Dr. Tian’s research projects were funded by ARO, NSF, DHS, Google, FXPAL, NEC, Blippar, SALSI, CIAS, Akiira Media Systems, HP and UTSA. He received 2017 UTSA President Distinguished Award for Research Achievement and 2010 Google Faculty Research Award. He is the Associate Editor of IEEE TMM, TCSVT, ACM TOMM, MMSJ, and MVA.
\end{IEEEbiography}

\begin{IEEEbiography}[{\includegraphics[width=1in,height=1.25in,clip,keepaspectratio]{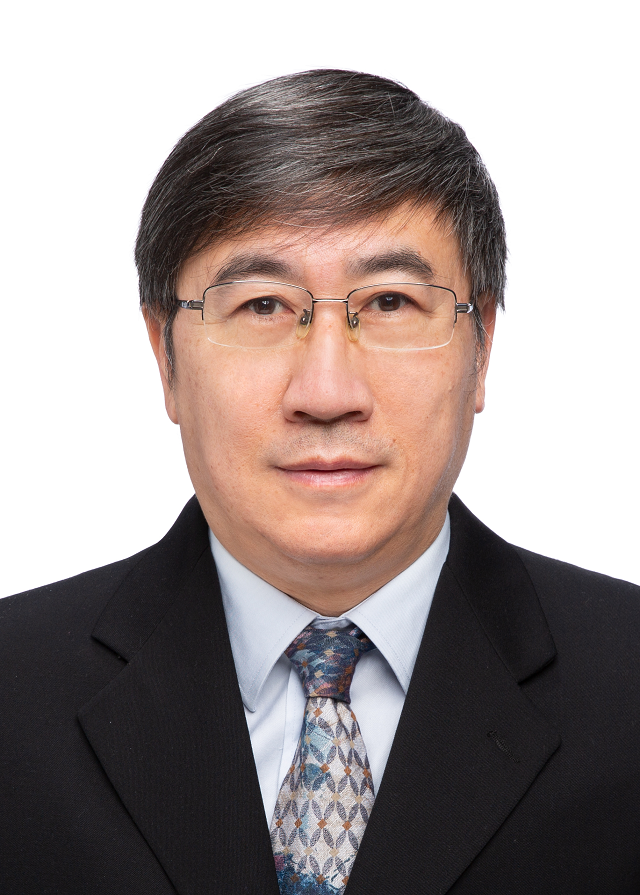}}]{Qingming Huang} is a chair professor in the University of Chinese Academy of Sciences and an adjunct research professor in the Institute of Computing Technology, Chinese Academy of Sciences. He graduated with a Bachelor degree in Computer Science in 1988 and Ph.D. degree in Computer Engineering in 1994, both from Harbin Institute of Technology, China. His research areas include multimedia computing, image processing, computer vision and pattern recognition. He has authored or coauthored more than 400 academic papers in prestigious international journals and top-level international conferences. He was the associate editor of IEEE Trans. on CSVT and Acta Automatica Sinica, and the reviewer of various international journals including IEEE Trans. on PAMI, IEEE Trans. on Image Processing, IEEE Trans. on Multimedia, etc. He is a Fellow of IEEE and has served as general chair, program chair, track chair and TPC member for various conferences, including ACM Multimedia, CVPR, ICCV, ICME, ICMR, PCM, BigMM, PSIVT, etc.
\end{IEEEbiography}




\end{document}